\pdfoutput=1  

\documentclass{article}

\usepackage{microtype}
\usepackage{graphicx}
\usepackage{booktabs}
\usepackage{hyperref}

\usepackage[accepted]{icml2026}

\usepackage{amsmath,amssymb,amsthm}
\usepackage{bm}
\usepackage{algorithm}
\usepackage{algorithmic}

\newcommand{\PEBS}{\textsc{PEBS}}

\newtheorem{theorem}{Theorem}
\newtheorem{proposition}{Proposition}

\icmltitlerunning{PEBS for RLHF Reward Calibration}

\begin{document}

\twocolumn[
  \icmltitle{\texorpdfstring{\PEBS{}}{PEBS}: Per-rater Empirical-Bayes
    Shrinkage for RLHF Reward-Model Calibration}

  \begin{icmlauthorlist}
    \icmlauthor{Arnav Raj}{iitd}
  \end{icmlauthorlist}

  \icmlaffiliation{iitd}{Department of Computer Science and Engineering,
    Indian Institute of Technology Delhi, New Delhi, India}

  \icmlcorrespondingauthor{Arnav Raj}{arnav.raj.cs522@cse.iitd.ac.in}

  \icmlkeywords{RLHF, pluralistic alignment, empirical Bayes,
    per-annotator calibration, reward modeling}

  \vskip 0.3in
]

\printAffiliationsAndNotice{}
\raggedbottom

\begin{abstract}
\noindent
Reward models for Reinforcement Learning from Human Feedback (RLHF)
pool preferences across thousands of annotators and fit one global
affine calibrator, collapsing raters with systematically different
rating-scale offsets and slopes into a single average-rater fit
that does not match any individual annotator. \PEBS{} is a per-rater empirical-Bayes
shrinkage estimator: it fits per-rater affine calibrators on a
held-out slice of each annotator's ratings and applies
Morris--James--Stein empirical-Bayes shrinkage toward the population
mean, in closed form and without retraining the reward model. On
PRISM, \PEBS{} reduces within-user held-out RMSE by
$\bm{8.58\%}$ over the pooled
population-slope baseline.
The procedure replicates on PluriHarms harm ratings (Qwen-2.5 base, in-family)
with a $\bm{+9.66\%}$ RMSE reduction over the same population-slope baseline.
PEBS is a closed-form post-hoc estimator for annotator-specific
affine calibration in RLHF reward modeling; it leaves the reward
base model unchanged and estimates only the rater-level map used at
inference time for new ratings.
\end{abstract}

\section{Introduction and Related Work}\label{sec:intro}

\begin{figure*}[t!]
\centering
\includegraphics[width=0.99\textwidth]{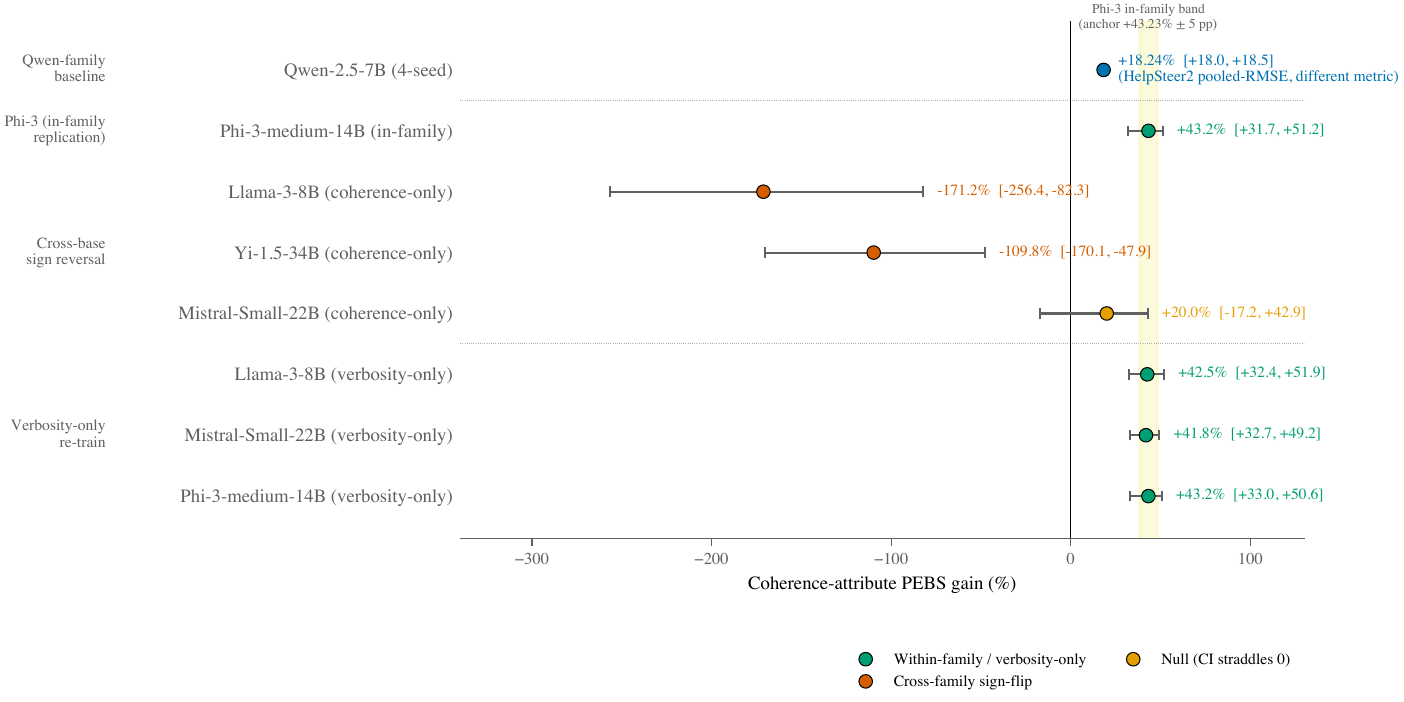}
\caption{\textbf{The Phi-3-medium-14B in-family case falls within
$\pm 5$\,pp of the single-seed anchor ($+43.23\%$, shaded band);
the Qwen-2.5 row replicates in-family on a different metric
(HelpSteer2 pooled-RMSE, $+18.24\%$).} Forest plot of
point estimates with $95\%$ row-cluster bootstrap confidence
intervals, grouped by base-model family. The three Llama-family-dense
bases are shown second as scope characterization: on a coherence head
they split into two negative outcomes and one wide-CI null. A
verbosity-only retrained head recovers a positive gain on the same
bases, pointing to a coherence-head/dense-architecture interaction
rather than an attribute-agnostic verbosity bias; calibration
diagnostics are in Appendix~\ref{app:adapter-inspection}.}
\label{fig:overview}
\end{figure*}

Reinforcement Learning from Human
Feedback~\citep{christiano2017deep,stiennon2020summarize,ouyang2022training}
assumes a Bradley--Terry~\citep{bradley1952rank} pairwise-preference
model: preferences from many annotators are pooled into one likelihood, a
scalar reward $r_\phi$ is fit, and the result is used for either
proximal-policy-optimization (PPO)-style RLHF or
DPO~\citep{rafailov2023direct}.\footnote{Code and fitted calibrators: \url{https://github.com/deadsmash07/pebs-pluralistic}.} The standard pooled-likelihood objective drops the annotator index $j$
from this aggregation,
which collapses raters with systematically different rating-scale
calibrations into a single global affine fit and confounds
calibration heterogeneity with reward signal.
Figure~\ref{fig:overview} previews the base-family transfer
summary: the procedure replicates on the Qwen-2.5 and Phi-3 reference rows,
turns negative on two of three Llama-family-dense
bases when trained on a coherence head, and recovers a positive
gain on those same bases under a verbosity-only run that
points to the coherence-head /
dense-architecture interaction (\S\ref{sec:crossfam}).

Different annotators use the $0$--$100$ score scale heterogeneously.
Some compress the scale, some stretch it, and some differ in
baseline. Pooling such observations naively yields a reward model
(RM) that fits the
\emph{average} rater, a fit that does not correspond to any
individual annotator. Several lines of work make the
measurement-validity problem explicit:
\citet{ghafouri2026measuring} argue that RLHF preference measurement
needs social-science diagnostics, and
\citet{ma2026personalizedrb} report that frontier RMs peak at
$75.9\%$ on \emph{their} per-user preference benchmark.
\citet{rmselectioncrisis2025} measure rank-correlation
$\tau{=}0.08$--$0.31$ (Kendall) between upstream RM pair-accuracy
and downstream policy accuracy on Pref-LaMP, a personalised-preference
benchmark. Together these indicate that a single global RM
degrades per-annotator accuracy even when its aggregate accuracy
is high.

Partial pooling is the classical fix, and per-annotator effect
modeling is the psychometric mainline outside RLHF. The Rasch model~\citep{rasch1960probabilistic} and classical
Item-Response Theory~\citep{baker2001basics} parametrize per-rater
difficulty and discrimination.
\citet{dawid1979maxlik} gave the canonical rater-effect mixture predating
modern crowdsourcing, and \citet{paun2018comparing} benchmark
hierarchical Bayesian rater models on NLP annotation, establishing
partial pooling as the dominant paradigm. In regression-style data
analysis, the textbook estimators are the Morris/James--Stein
empirical-Bayes (EB) shrinkage~\citep{robbins1956empirical,morris1983parametric}
and the
Best Linear Unbiased Predictor (BLUP)~\citep{henderson1975best},
the canonical EB estimator from linear-mixed-model theory, and
the blending weight $\omega = \tau^2/(\tau^2 + V)$ is standard
in hierarchical modeling~\citep{gelman2007data,pinheiro2000mixed}.

\paragraph{Per-user reward modeling in RLHF.}
The pluralistic-alignment programme outlined by
\citet{sorensen2024roadmap} distinguishes Overton, steerable,
and distributional axes (with \citet{bakker2022fineagree}
establishing direct upstream evidence on language-model
fine-tuning toward per-annotator agreement);
\citet{conitzer2024socialchoice} argue that
aggregating diverging human feedback is a social-choice problem;
benchmarks and datasets in this line include
\citet{castricato2025persona} (PERSONA, persona-conditioned
preferences) and \citet{zhang2025community} (Community Alignment,
multilingual representative-sample preferences with
negatively-correlated candidate sampling). The label \PEBS{} denotes
the per-rater empirical-Bayes shrinkage estimator used here:
operationally, it shrinks annotator-specific affine calibration
parameters. The method
operates on a complementary axis (see \S\ref{sec:discussion}):
per-annotator calibration heterogeneity. RLHF work has also explored
per-user effects along several distinct estimator axes.
\citet{kobalczyk2025preference} formulate
user-specific factor confounding in a causal framework for preference
learning.
\citet{pgenrm2026} use learned user prototypes; PEBS instead uses
stable per-rater identifiers. \citet{liu2025uq} model rater
rationality as a function of annotator context. Whether demographic
covariates suffice for the per-user effect is testable: an
analysis-of-variance (ANOVA, partitioning between- versus
within-group variance) of the fitted per-user calibrators against
six PRISM annotator features (age, gender, region, education,
political orientation, English fluency; \S\ref{sec:stress-tests}) leaves only the
gender-to-$\hat\beta_j$ effect surviving Bonferroni correction at
$\eta^2{=}0.018$ (here $\hat\beta_j$ is the per-rater offset
estimator from Section~\ref{sec:method}), so demographic grouping cannot substitute for
per-user calibration. The most closely related empirical-Bayes
shrinkage method, \textsc{EBPO}~\citep{han2026ebpo}, shrinks
per-prompt group-relative-policy-optimization (GRPO) advantage
baselines on verifiable-reward tasks,
which targets a different scale (per-prompt advantage, not
per-rater calibration). A comparison of PEBS against
these related methods appears in Table~\ref{tab:method-axis}
(appendix).

\paragraph{Contributions.}
First, \PEBS{} puts a classical correction where RLHF reward
pipelines usually omit it: Efron--Morris--James--Stein partial
pooling~\citep{efron1973stein} for annotator-specific scale and
offset, applied post hoc to scalar RM outputs. Under annotator
heterogeneity, this correction materially helps calibration-sensitive
losses. The result in this setting is a within-user RMSE reduction of
$8.58\%$ on PRISM with a Qwen-2.5-7B base model
(Table~\ref{tab:t1-within}); the procedure replicates on PluriHarms
harm ratings (${+}9.66\%$; \S\ref{sec:4corpora}) and on a
same-family Phi-3-medium-14B reference ($+42.15\%$ across five
seeds, all positive; \S\ref{sec:crossfam}). The estimator is
closed-form and operates downstream of any reward model's scalar
predictions; an ablation (\S\ref{sec:nulls}) separates the gain
into a textbook Efron--Morris intercept-shrinkage floor, which
appears even under a signal-free (permuted) reward, and a smaller
\PEBS{}-specific slope-shrinkage residual that requires real reward
signal. A pre-registered four-base coherence-only probe
(\S\ref{sec:crossfam}, Table~\ref{tab:base-transfer-grid})
identifies the transfer limit structurally: the procedure transfers
within the Qwen-2.5 family and on the Phi-3-medium-14B reference,
while under coherence-only training on Llama-family-dense bases two
of three turn negative; a paired verbosity-only control recovers
positive gain on the same bases, pointing to a coherence-head /
dense-architecture interaction rather than attribute-agnostic
verbosity bias. We report this scope boundary without claiming
generality. On the theory side we prove (\S\ref{sec:oracle},
Theorem~\ref{thm:oracle}) that a sample-split variant of PEBS's
slope shrinkage stays within a $(1+c/J)$ factor of an oracle that
knows the true slope variance, with an explicit constant; a
PRISM-calibrated simulation of the deployed estimator puts the
realized risk inflation near $0.2\%$. A closed-form Morris
$g$-function forecaster (\S\ref{sec:morrisg}) predicts PEBS gain
on a new corpus from a short pilot, validated to within $0.2$\,pp
on four rating corpora. Table~\ref{tab:method-axis} (appendix)
contrasts these extensions of the Efron--Morris--James--Stein
estimator~\citep{efron1973stein,morris1983parametric,henderson1975best}
with the most closely related personalization methods.

\section{Method}\label{sec:method}

\subsection{Partial-pooling estimator}\label{sec:pp}
Given observations $\{y_{ji}\}$ indexed by annotator $j$ and utterance
$i$, the complete-pooling estimator ignores $j$. We instead estimate a
cluster-specific parameter $\theta_j$ via the classical empirical-Bayes
blend
\begin{align}
  \hat\theta_j^{\mathrm{PP}}
  &\;=\;
  \omega_j\,\hat\theta_j^{\mathrm{local}}
  + (1-\omega_j)\,\hat\theta_{\mathrm{pool}}, \\
  \omega_j &\;=\; \frac{\tau^2}{\tau^2 + V(\hat\theta_j^{\mathrm{local}})},
  \label{eq:shrinkage}
\end{align}
where $\tau^2$ is the cross-cluster variance of $\theta_j$ and
$V(\hat\theta_j^{\mathrm{local}})$ is the within-cluster sampling
variance. Eq.~\eqref{eq:shrinkage} is the Morris/James--Stein
empirical-Bayes shrinkage~\citep{morris1983parametric} and recovers the
BLUP of the linear mixed model~\citep{henderson1975best}. At $\omega{=}0$
it reduces to the pooled estimator and at $\omega{\to}1$ it reduces
to per-cluster OLS, with the closed-form $\omega_j(n_j)$ curve and
small-$n_j$ down-weighting visualized in Appendix
Figure~\ref{fig:omega-shrinkage}.

\subsection{Per-user calibration model}\label{sec:t1-method}
Algorithm~\ref{alg:pebs} sets out the three-stage procedure (shared
reward model; per-rater OLS calibrator; EB shrinkage) end-to-end.
For each annotator $j$ and utterance $i$, we model the user's continuous
preference score as
\begin{equation}
  s_{ji} \;=\; \alpha_j\,\hat r_\phi(x_{ji}) \;+\; \beta_j \;+\; \varepsilon_{ji},
  \label{eq:linear-model}
\end{equation}
where $\hat r_\phi$ is a shared reward model fine-tuned on pooled PRISM
preferences and $(\alpha_j,\beta_j)$ is a per-user linear calibrator:
$\alpha_j$ is the per-annotator multiplicative slope (the units in
which annotator $j$ converts a unit of model reward into a unit of
self-reported score) and $\beta_j$ is the per-annotator additive offset
(the baseline score $j$ assigns to a zero-reward response). Per-user OLS yields
$\hat\alpha_j^{\mathrm{OLS}},\hat\beta_j^{\mathrm{OLS}}$ with sampling
variance $V(\hat\alpha_j) = \hat\sigma_\varepsilon^2/\bigl(n_j\,\mathrm{Var}_j(\hat r_\phi(x_{ji}))\bigr)$.
The EB-shrunk estimator is the direct application of
Eq.~\eqref{eq:shrinkage}:
\begin{equation}
  \hat\alpha_j^{\mathrm{shrunk}}
  \;=\;
  \omega_\alpha^{(j)}\,\hat\alpha_j^{\mathrm{OLS}}
  \;+\;(1-\omega_\alpha^{(j)})\,\alpha_{\mathrm{pop}},
\end{equation}
with $\omega_\alpha^{(j)} = \hat\tau_\alpha^2 / (\hat\tau_\alpha^2 + V(\hat\alpha_j))$
and an analogous formula for $\hat\beta_j^{\mathrm{shrunk}}$.
$\hat\tau_\alpha^2$ is a Method-of-Moments (MoM) estimate on the
per-user $\hat\alpha_j^{\mathrm{OLS}}$ distribution; a Restricted
Maximum Likelihood (REML) cross-check on the two-level
(rater, observation) mixed model
$s_{ji} \sim \beta_j + \alpha_j\hat r_\phi(x_{ji}) +
\varepsilon$~\citep{seabold2010statsmodels,pinheiro2000mixed} disagrees
on PRISM by $3.5\%$ on $\hat\tau_\alpha^2$ and $11.1\%$ on
$\hat\tau_\beta^2$; since the EB risk is stationary in $\tau^2$ at the
truth (\S\ref{sec:oracle}, Appendix~\ref{app:proof-oracle}, Step~2), a
few-percent error in $\hat\tau^2$ perturbs the risk only at second
order. The fitted cross-user correlation
between $\hat\alpha_j$ and $\hat\beta_j$ is small (point estimate
$0.09$), which supports the separable EB shrinkage in
Algorithm~\ref{alg:pebs}: the per-user slope $\alpha_j$ and offset
$\beta_j$ can be shrunk independently rather than jointly with a
$2\!\times\!2$ covariance matrix.

\begin{algorithm}[t]
\caption{\PEBS{}: per-rater empirical-Bayes shrinkage}
\label{alg:pebs}
\begin{algorithmic}[1]
\STATE \textbf{Input:} reward model $\hat r_\phi$;
  per-rater calibration set $\{(x_{ji},s_{ji})\}_{j,i}$, where $x_{ji}$
  is the $i$-th utterance from rater $j$ and $s_{ji}\in[0,100]$ is the
  rated score; the per-user covariate is the RM prediction $\hat r_\phi(x_{ji})$.
\FOR{each rater $j$ with $n_j{\ge}3$ \COMMENT{PRISM uses $n_j{\ge}6$, \S\ref{sec:setup}}}
  \STATE $(\hat\alpha_j^{\mathrm{OLS}},\hat\beta_j^{\mathrm{OLS}})
       \!\leftarrow\! \mathrm{OLS}(\hat r_\phi(x_{j\cdot}),\, s_{j\cdot})$;
    $V(\hat\alpha_j) = \hat\sigma_\varepsilon^2 / \bigl(n_j\,\mathrm{Var}_j(\hat r_\phi(x_{ji}))\bigr)$
\ENDFOR
\STATE MoM:
  $\hat\tau_\alpha^2\!\leftarrow\!\mathrm{Var}_j(\hat\alpha_j^{\mathrm{OLS}})
   - \overline{V(\hat\alpha_j)}$
\STATE Truncate: $\hat\tau_\alpha^2 \leftarrow \max(0,\,\hat\tau_\alpha^2)$ \quad \COMMENT{standard EB truncation, \citet{morris1983parametric} \S4}
\STATE Per-rater weights: $w_j = 1/V(\hat\alpha_j)$
\STATE Population mean: $\alpha_{\mathrm{pop}} = \big(\sum_j w_j \hat\alpha_j^{\mathrm{OLS}}\big) \big/ \big(\sum_j w_j\big)$
\FOR{each rater $j$}
  \STATE Weight:
    $\omega_\alpha^{(j)}\!\leftarrow\!\hat\tau_\alpha^2/
       (\hat\tau_\alpha^2 + V(\hat\alpha_j))$
  \STATE Shrunk:
    $\hat\alpha_j^{\mathrm{shrunk}} \leftarrow
    \omega_\alpha^{(j)}\,\hat\alpha_j^{\mathrm{OLS}}
    + (1{-}\omega_\alpha^{(j)})\,\alpha_{\mathrm{pop}}$
  \STATE Analogously for $\hat\beta_j^{\mathrm{shrunk}}$ (with $\hat\tau_\beta^2$ truncated at zero).
\ENDFOR
\STATE \textbf{return} $\{(\hat\alpha_j^{\mathrm{shrunk}},\hat\beta_j^{\mathrm{shrunk}})\}_{j=1}^{J}$
\end{algorithmic}
\end{algorithm}

\begin{proposition}[Pair-accuracy invariance under \PEBS{}]\label{prop:mi}
Assume $\alpha_{\mathrm{pop}} > 0$ and that the post-shrinkage slopes
are strictly positive,
$\hat\alpha_j^{\mathrm{shrunk}} > 0$ for every rater $j$. Then the
affine map $r \mapsto \hat\alpha_j^{\mathrm{shrunk}} r +
\hat\beta_j^{\mathrm{shrunk}}$ is strictly monotone and preserves
the argmax of every finite list, so any pair-accuracy or
best-of-$n$ benchmark is constant across the pop-slope and
EB-shrunk arms; gains can only appear in calibration-sensitive
losses such as root mean squared error (RMSE) and the
Bradley--Terry negative log-likelihood (NLL).
\end{proposition}
\noindent\emph{(Proof: monotonicity.)} The positivity assumption is
not automatic: since $\hat\alpha_j^{\mathrm{shrunk}}$ is a convex
combination of $\hat\alpha_j^{\mathrm{OLS}}$ and
$\alpha_{\mathrm{pop}}$, a sufficiently negative per-rater OLS slope
can produce a negative shrunk slope whenever
$\omega_\alpha^{(j)}>0$; it holds automatically only in the
fully-pooled case $\hat\tau_\alpha^2 = 0$. We therefore verify it
empirically: one of $1{,}394$ raters has a marginally negative shrunk
slope on PRISM (minimum $-0.33$); the measured pair accuracy is
nonetheless identical across the pop-slope and EB-shrunk arms
($0.6834$ both, \S\ref{sec:nulls}), so the invariance holds
exactly on the evaluated cohort.
\noindent\emph{Consequence:} the held-out pair-accuracy null
reported in \S\ref{sec:nulls} is required rather than
disconfirming, and PEBS is orthogonal to argmax-style
benchmarks such as
RewardBench~2~\citep{lambert2024rewardbench}.

\subsection{PRISM setup and base reward model}\label{sec:setup}
We use the PRISM Alignment corpus~\citep{kirk2024prism}, a public
RLHF dataset that exposes stable per-annotator IDs alongside multi-turn
preference judgments at the scale we require. Two nested cohorts
enter the paper. The reward model is trained on $26{,}876$ preference
pairs from the $1{,}391$ demographic-complete participants
(${\sim}93\%$ of PRISM's $1{,}500$; $75$ countries, $24$ demographic
axes), under a stratified $80/20$ held-out-user split ($21{,}474$
train / $5{,}402$ test pairs, $1{,}113$ train / $278$ held-out users,
no within-user leakage). The per-rater calibrators are fit on
utterance-level scores with an $n_j{\ge}6$-observation filter, which
retains $J{=}1{,}394$ of $1{,}396$ extractable participants; all
within-user calibration results (\S\ref{sec:withinuser} onward)
use this $1{,}394$-user cohort.

The base reward model $\hat r_\phi$ is Qwen2.5-7B-Instruct~\citep{qwen25}
fine-tuned with low-rank adaptation~\citep{hu2022lora} ($r{=}32$) on the
pooled PRISM preferences with the centered-rewards regularizer of
\citet{eisenstein2024helping}; full training-loop configuration is in
Appendix~\ref{app:replication}. The base model reaches $64.00\%$ pair
accuracy on the held-out-user test set, roughly three percentage points
over a matched Qwen2.5-0.5B baseline. All PEBS calibrators are fit
on this 7B base model's scores. Code, configurations, and
fitted per-rater calibrator weights are in the public repository
linked from the footnote in \S\ref{sec:intro}.

The HelpSteer2 across-family probes (\S\ref{sec:crossfam}) train
attribute-specific reward heads: the coherence head trains the LoRA
adapter to predict per-row HelpSteer2 coherence scores, and the
verbosity head substitutes verbosity scores under an otherwise
identical training configuration. The verbosity-only run is a control that
tests whether a negative outcome on the coherence head reflects a
coherence-specific phenomenon or an attribute-agnostic upstream-bias
effect.

\section{Experiments}\label{sec:exp}

We evaluate PEBS along three axes:
\textbf{(a) within-user calibration accuracy} on PRISM
(\S\ref{sec:withinuser}--\S\ref{sec:effectsize}: RMSE, paired effect
size, Bradley--Terry NLL),
\textbf{(b) cross-corpus replication} (\S\ref{sec:4corpora}:
PluriHarms on a Qwen-2.5 base model; HelpSteer2 multi-attribute
observation in Appendix~\ref{app:helpsteer-multi-attribute}), and
\textbf{(c) base-family transfer} (\S\ref{sec:crossfam}: a
pre-registered four-base scope panel with a verbosity-only
control). Two theoretical tools frame these empirics: an
oracle inequality for slope shrinkage (\S\ref{sec:oracle}) and a
Morris $g$-function closed-form forecaster
(\S\ref{sec:morrisg}). Stress tests
(\S\ref{sec:stress-tests}) and pre-registered ablations
(\S\ref{sec:nulls}) follow.

\subsection{Held-out PRISM prediction}\label{sec:withinuser}
Table~\ref{tab:t1-within} reports four-arm performance on $N{=}1{,}394$
users with $k{=}5$-fold cross-validation (CV). The EB-shrunk
calibrator of Eq.~\eqref{eq:shrinkage} yields an $\bm{8.58\%}$
relative within-user RMSE reduction over the pop-slope baseline
(a single global affine calibrator
$(\alpha_{\mathrm{pop}}, \beta_{\mathrm{pop}})$ fit by pooled OLS,
the strongest of the no-personalization arms we evaluate). Naive per-user OLS is rarely used in practice: although the
regression is computationally negligible, each
per-user fit overfits its own small sample~$n_j$: low-$n_j$ users
get high-variance calibrators and held-out RMSE worsens. Shrinking
each per-user fit toward the population mean by the closed-form
weight of Eq.~\eqref{eq:shrinkage} closes that gap at near-zero
marginal cost.

\begin{table}[htbp]
\caption{\textbf{PEBS recovers within-user RMSE on PRISM beyond
what naive per-user OLS achieves.} Held-out score-prediction
RMSE for $N{=}1{,}394$ users with $k{=}5$ cross-validation on a 7B
base model. The EB-shrunk estimator dominates naive per-user OLS on
$77.3\%$ of users (sign test, $p{<}10^{-92}$).}\label{tab:t1-within}
\centering
\footnotesize
\setlength{\tabcolsep}{2pt}
\begin{tabular}{lcccc}
\toprule
Arm & RMSE${\downarrow}$ & Med.${\downarrow}$ & $\Delta$ vs pop${\downarrow}$ & Wilcoxon $p$ \\
\midrule
No calibration    & $27.13$      & $27.13$ & $+6.3\%$        & --                      \\
Population slope  & $25.52$      & $25.26$ & baseline        & --                      \\
Per-user OLS      & $23.73$      & $23.36$ & $-7.02\%$       & $1.2{\times}10^{-64}$   \\
EB-shrunk (ours)  & $\bm{23.33}$ & $23.12$ & $\bm{-8.58\%}$  & $\bm{3.8{\times}10^{-108}}$ \\
\bottomrule
\end{tabular}
\end{table}

Using $4{,}000$-replicate cluster bootstrap by user~\citep{cameron2008bootstrap,efron1987better},
the bias-corrected accelerated (BCa) 95\% CI on the
$8.58\%$ relative gain is $[7.59\%, 9.42\%]$, excluding zero.

\subsection{Effect size and BT log-likelihood}\label{sec:effectsize}

On the same PRISM cohort the per-user paired effect of the RMSE
drop (mean per-user difference between EB-shrunk and pop-slope arms
divided by the within-user paired-difference SD) is
$\bm{d_{\text{paired}}{=}0.542}$
($95\%$ CI $[0.491,\,0.607]$), roughly half the within-user
re-rating noise on the $0$--$100$ scale. The cross-user pooled
reduction is $0.075$ SD; the two readings differ because they
condition on within-user vs. marginal variance, and the per-user
calibrator targets the within-user component. Within-user RMSE is
a proxy for downstream reward-model behaviour; the quantity that
enters the RLHF reward-model loss directly is the held-out pairwise
Bradley--Terry (BT) negative log-likelihood (NLL), which is not
monotone-invariant in the calibrator (unlike pair accuracy). On the
held-out preference pairs the mean per-pair BT-NLL improves by
$\bm{5.7\%}$ relative (paired-$t$ $p{<}10^{-7}$). The improvement
is tail-concentrated rather than uniform: the per-pair Wilcoxon $p$
is $0.77$ and the median $\Delta\mathrm{NLL}$ is near zero, with
the gain carried by a minority of users with atypical $\beta_j$ on
hard pairs.

\subsection{Downstream calibration losses}\label{sec:why-calib}
DPO and PPO consume reward scores as scalars, not as ranks. The
DPO loss
$-\log\sigma(\beta\,(r_{\text{chosen}}{-}r_{\text{rejected}}))$
\citep{rafailov2023direct} is a sigmoid of a magnitude difference.
PPO advantage normalization operates on raw scores
\citep{schulman2017proximal,ouyang2022training}. BT-NLL weights
each preference pair by the magnitude of its score gap. By
Proposition~\ref{prop:mi}, PEBS does not change pair accuracy;
however, it changes the gradient that the policy training step uses.
Multi-attribute aggregation compounds the issue: per-rater scale
heterogeneity distorts the sum of raw scalars. Reward-model overoptimization is the
limiting failure-mode of poor downstream calibration
\citep{gao2023scaling}; a pre-registered PPO probe on PRISM
(Qwen-2.5-7B policy with the Skywork-Llama-3.1-8B reward
model~\citep{liu2024skywork})
shows the uncorrected reward collapses at
$\mathrm{KL}{\ge}1.0$ while the \PEBS{}-shrunk arm holds
(judge-reward gap ${+}2.16$, conservative $95\%$ CI excluding
zero). The RM-selection literature reports
upstream-vs-downstream rank-correlations of only
$\tau{=}0.08$--$0.31$~\citep{rmselectioncrisis2025}. \PEBS{}
targets calibration-sensitive losses; improving pair accuracy
requires a separate selection-style component
(\S\ref{sec:discussion}).

\subsection{Cross-corpus replication}\label{sec:4corpora}
\begin{figure}[htbp]
\centering
\includegraphics[width=0.9\columnwidth]{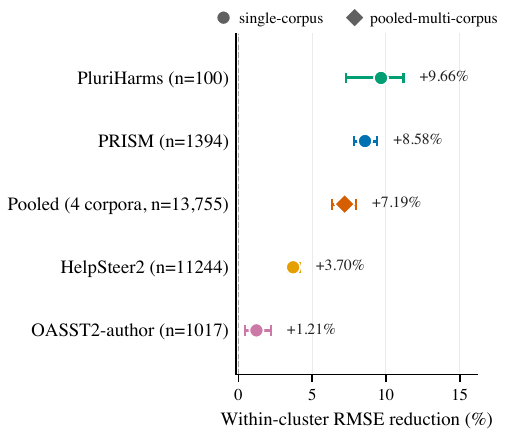}
\caption{\textbf{PEBS reduces RMSE on four single-corpus
replications and on a $195{,}963$-observation pooled corpus, all
using a single Qwen-2.5 base model.} Horizontal forest of
within-cluster gain
(\%) with $95\%$ BCa cluster-bootstrap CIs; circles are
single-corpus replications, the diamond is the four-corpus pooled
estimate. The dashed reference at zero is the pop-slope baseline. The
pooled-multi-corpus row ($+7.19\%$ $[+6.36, +7.96]$) uses
namespaced cluster IDs across the four corpora.}
\label{fig:cross-corpus-forest}
\end{figure}

A single-corpus result on PRISM does not by itself establish a
pluralism claim. We replicate the within-cluster-RMSE evaluation on
three additional corpora with stable cluster IDs (PluriHarms harm
ratings~\citep{li2026pluriharms}, whose taxonomy follows the
value-annotation tradition of KALEIDO~\citep{sorensen2024kaleido};
HelpSteer2 prompt-cluster
attributes~\citep{wang2024helpsteer2}; OASST2
authors~\citep{kopf2023openassistant}) and on a single
heterogeneous-cluster pool of all four
($195{,}963$ observations, $13{,}755$ namespaced clusters,
per-corpus $z$-score normalization). All five rows reduce RMSE on
the same Qwen-2.5 base model
(Figure~\ref{fig:cross-corpus-forest}); OASST2-author is the weakest
replication (${+}1.21\%$; its bootstrap CI excludes zero though a
per-cluster Wilcoxon test does not reach significance); PluriHarms ($+9.66\%$) and
PRISM ($+8.58\%$) agree to within $\sim 1$\,pp of each other despite measuring
harm ratings versus preferences, consistent with (though not establishing)
a cluster-scale and not feedback-type-specific mechanism. The HelpSteer2
row treats prompt-cluster attribute scores as the cluster axis (a
different problem-geometry from per-annotator pluralism); the
per-attribute breakdown is in
Appendix~\ref{app:helpsteer-multi-attribute}. An ordinal
preference corpus (MultiPref) lies outside the Gaussian-RE
scope and is documented separately in \S\ref{sec:morrisg}.

\subsection{Cross-family transfer}\label{sec:crossfam}

A pre-registered four-base coherence-only probe (Meta-Llama-3-8B,
Mistral-Small-22B, Yi-1.5-34B, Phi-3-medium-14B; two
mixture-of-experts (MoE) runs, Phi-3.5-MoE and Mixtral-8$\times$7B,
are reported as appendix-only
boundary evidence in App.~\ref{app:adapter-inspection}) with five
training seeds on the same-family Phi-3 reference and a paired
verbosity-only run on the three Llama-family-dense bases together
map where the HelpSteer2 multi-attribute observation
(Appendix~\ref{app:helpsteer-multi-attribute}) transfers beyond
Qwen-2.5; Table~\ref{tab:base-transfer-grid} summarizes the result.
The pre-registered sign-flip criterion is met for Llama-3-8B and Yi-1.5-34B;
Mistral-Small-22B is a single-seed null; Phi-3-medium-14B holds at
$+42.15\%$ across $5$ seeds (positive in all five).
Both columns of Table~\ref{tab:base-transfer-grid} report the
held-out \emph{coherence-attribute} gain; the columns differ in
which attribute the head was trained on. Under verbosity-only
training the untrained coherence head remains positive across all
four bases (single-seed for the three Llama-family-dense bases,
five-seed for Phi-3), while the trained verbosity head itself turns
negative (e.g.\ $-32.62\%$ on Phi-3;
Appendix~\ref{app:adapter-inspection}). This is evidence against
attribute-agnostic verbosity bias as the source of the
coherence-head reversal. A within-Llama
intervention sweep (zero-out, scramble, signal-content
replacement; two seeds each) further refines the mechanism: only
information-removal interventions reproduce the negative outcome, while
signal substitution preserves the same-family positive,
consistent with a collapse-by-removal pattern at within-Llama
scope. The full HelpSteer2 five-attribute breakdown,
calibration-diagnostic signatures, and the MoE-branch
partial-coverage LoRA scope are in
Appendix~\ref{app:adapter-inspection}.

\begin{table}[htbp]
\caption{\textbf{The verbosity-only control preserves the coherence
head on the three Llama-family-dense bases.} Each cell is the
held-out coherence-attribute gain (\%); columns differ in the
attribute the head was trained on. Phi-3-medium-14B is the
same-family five-seed reference (mean $+42.15\%$, Student-$t$
$95\%$ CI $[+40.10, +44.20]$); under verbosity-only training the
untrained coherence head stays positive on all four bases.}
\label{tab:base-transfer-grid}
\centering\scriptsize
\setlength{\tabcolsep}{2.5pt}
\begin{tabular}{lcc}
\toprule
 & \multicolumn{2}{c}{Coherence gain by trained attribute} \\
\cmidrule(lr){2-3}
Base & Coherence-trained & Verbosity-trained \\
\midrule
Phi-3-medium-14B & $+42.15\%$ & $+43.18\%$ \\
Llama-3-8B              & $-171.16\%$ & $+42.53\%$ \\
Mistral-Small-22B       & $+19.99\%$  & $+41.83\%$ \\
Yi-1.5-34B              & $-109.76\%$ & $+43.05\%$ \\
\bottomrule
\end{tabular}
\end{table}

\subsection{Oracle inequality for EB slope-shrinkage}\label{sec:oracle}
Beyond the empirical PRISM gain on Qwen-family base models, PEBS admits
an oracle inequality under the random-effects assumptions stated below.
Write $V_j$ for the within-annotator sampling variance of
$\hat\alpha_j^{\mathrm{OLS}}$ (\S\ref{sec:t1-method}) and
$M = \max_j V_j/\tau^2_\alpha$ for the noise-to-signal bound;
$R_{\text{oracle}}$ is the squared-error risk of the oracle estimator
that knows the true $\tau^2_\alpha$ and $R_{\text{EB}}$ that of the
truncated Morris MoM EB estimator.
\begin{theorem}[EB slope-shrinkage oracle inequality]\label{thm:oracle}
Let $J\ge 4$ denote the number of annotators. Assume the random-effect
DGP $\alpha_j=\alpha_{\mathrm{pop}}+u_j$ with
$u_j\sim\mathcal N(0,\tau^2)$ i.i.d.\ across $j$;
$\hat\alpha_j^{\mathrm{OLS}}\mid\alpha_j \sim \mathcal N(\alpha_j,V_j)$
independent across $j$, with $V_j$ and $\alpha_{\mathrm{pop}}$ known;
and $V_j \le M\tau^2$ for all $j$. Let $\hat\tau^2$ be the truncated
method-of-moments estimate computed on an auxiliary set of raters
drawn from the same DGP, independent of the $J$ raters being estimated
(sample splitting). Then
\begin{align}
  R_{\text{EB}} \le{}& \Bigl(1+\frac{c}{J}\Bigr) R_{\text{oracle}}
  \notag \\
  &+ 2\max(1,M)\,\tau^2 \exp\!\Bigl(-\tfrac{c_2 (J-1)}{(1+M)^2}\Bigr),
  \label{eq:oracle-ineq}
\end{align}
with $c \le \tfrac{64}{3}(1+M)^2$ and $c_2>0$ an absolute constant.
\end{theorem}
The proof (Appendix~\ref{app:proof-oracle}) adapts the
heteroskedastic-location analysis of \citet{xie2012sure} to the
OLS-slope statistic: the oracle is a stationary point of the
per-rater risk, so the $\hat\tau^2$ estimation error enters only at
second order. The constant is a conservative worst-case bound driven
by the sparsest raters (on PRISM, $n_j$ spans $6$--$144$, giving
$M\approx 4$); the deployed estimator additionally estimates
$\hat\tau^2$ and $\alpha_{\mathrm{pop}}$ on the same sample, a
coupling Appendix~\ref{app:proof-oracle} scopes and validates by
simulation. \emph{Operational consequence:} a PRISM-calibrated
simulation of the deployed estimator ($J{=}1{,}394$, $100$ seeds)
puts the expectation-level risk inflation at ${\approx}0.2\%$ (mean
risk ratio $1.002$; worst seed $1.017$), far too small to explain
the $8.58\%$ PRISM gain.

\subsection{Morris $g$-function forecaster}\label{sec:morrisg}
Given $(\tau^2_\alpha, \tau^2_\beta, \sigma^2_\varepsilon, \{n_j\},
\{\mathrm{Var}_{\mathrm{w}}(x_j)\})$ from a short pilot, where $x$
here denotes the RM score centered within each rater's calibration
slice ($\bar x_j = 0$, as in our implementation) and
$\mathrm{Var}_{\mathrm{w}}(x_j)$ its within-rater variance, the
two-parameter Morris risk-gap formula
\begin{align}
  \mathbb E\!\left[R_{\text{POP}} - R_{\text{EB}}\right]
  &= \tfrac{1}{J}\textstyle\sum_{j}\Bigl[\tau^2_\alpha
     \mathrm{Var}_{\mathrm{w}}(x_j)\,g\bigl(r_\alpha^{(j)}\bigr)\Bigr.
     \notag \\
  &\qquad \Bigl.{+}\ \tau^2_\beta\,g\bigl(r_\beta^{(j)}\bigr)\Bigr],
  \quad g(r) = r/(1+r),
  \label{eq:morrisg-2p}
\end{align}
with $r_\alpha^{(j)} = n_j\tau^2_\alpha\,\mathrm{Var}_{\mathrm{w}}(x_j)/\sigma^2_\varepsilon$ and
$r_\beta^{(j)} = n_j\tau^2_\beta/\sigma^2_\varepsilon$
(per-rater centering makes the slope and offset shrinkage gaps
separate; the predicted risk gap converts to the reported relative
RMSE reduction through division by $R_{\text{POP}}$),
predicts PEBS gain on a new corpus before running the full
estimation procedure. In practice, one estimates
$(\tau^2,\sigma^2,\{n_j\})$ on a short pilot, plugs into
Eq.~\eqref{eq:morrisg-2p}, and decides whether fitting the full PEBS
estimator is warranted. Table~\ref{tab:morrisg-forecast} validates the
forecast within $0.2$\,pp on the four continuous-rating corpora.
The MultiPref row shows the scope limit: the forecaster predicts a
large gap when the ordinal preference setting violates the
Gaussian random-effects assumption.

\begin{table}[htbp]
\caption{\textbf{The closed-form Morris $g$-function forecasts
\PEBS{} gain from a short pilot to within $0.2$\,pp on the four
continuous-rating corpora.} Observed gains use a leave-one-row-out
per-cluster CV matched to the forecaster's assumptions; the
OASST2-author row therefore differs by protocol, covariate, and
cohort from the \S\ref{sec:4corpora} replication row
(${+}1.21\%$), and the two are not comparable. At SHP's cluster
sizes $\omega\!\to\!1$, where PEBS reduces to per-cluster OLS and
the exact forecast match is expected rather than informative.
MultiPref is the ordinal-preference
limit: its predicted-versus-observed gap flags a Gaussian
random-effects mismatch.}
\label{tab:morrisg-forecast}
\centering\scriptsize
\setlength{\tabcolsep}{2.5pt}
\begin{tabular}{lcccc}
\toprule
Corpus & pred.\ (2-p) $\uparrow$ & observed $\uparrow$ & $|\Delta|$ $\downarrow$ & setting \\
\midrule
PRISM             & $8.43\%$ & $8.58\%$ & $0.15$~pp & finite-$r$ \\
PluriHarms        & $8.81\%$ & $8.64\%$ & $0.17$~pp & finite-$r$ \\
OASST2-author     & $8.37\%$ & $8.33\%$ & $0.04$~pp & finite-$r$ \\
SHP-subreddit     & $8.00\%$ & $8.00\%$ & $0.00$~pp & $\omega\!\to\!1$ \\
MultiPref & $17.96\%$ & $0.47\%$ & $17.49$~pp & ord.\ misspec. \\
\bottomrule
\end{tabular}
\end{table}

The MultiPref row in Table~\ref{tab:morrisg-forecast} is the
calibrated null: an ordinal preference
corpus~\citep{miranda2024hybrid} on which the
forecaster's $17.49$\,pp predicted-versus-observed gap correctly
flags Gaussian-RE mis-specification (\S\ref{sec:limitations}).

\subsection{Stress tests}\label{sec:stress-tests}

A natural concern is that $k{=}5$ random-fold CV may overstate
deployment generalization if the per-user $(\alpha_j,\beta_j)$
parameters are not time-invariant. Across five pre-registered
seeds the random-fold gain is tight around the $8.58\%$
point estimate: all five seed CIs exclude zero and the per-seed
gains lie within $0.17$\,pp of it. We repeated the within-user
evaluation with a strict temporal $80/20$ split, sorting utterances
by PRISM generation timestamp. The shrinkage gain holds at
$\bm{7.55\%}$, with a $30$-seed cluster-bootstrap CI that brackets
the random-CV point estimate, so the within-user RMSE
result also holds under a stricter temporal split. Across three
base reward models (Qwen2.5-7B,
Skywork-Reward-Gemma-2-27B~\citep{liu2024skywork},
Llama-3.2-3B-Instruct) crossed with PRISM and PluriHarms, all six
cells return a positive shrinkage gain whose $95\%$ CI strictly
excludes zero, even though the HelpSteer2 multi-attribute
observation does not extend across architectures
(\S\ref{sec:crossfam},
Appendix~\ref{app:helpsteer-multi-attribute}). The PRISM gain is
also stable across thirty-four subsets covering
top-$|\hat\alpha_j|$ trimming, small-and-large-$n$ slices, random
user subsamples, and demographic cells, and demographic grouping
cannot replace per-user calibration on PRISM (only
gender${\to}\hat\beta_j$ survives Bonferroni at small explained
variance $\eta^2{<}0.02$; see Appendix~\ref{app:prereg-mechanics}
for the six-demographic ANOVA detail).

\paragraph{Cold-start threshold.}
EB shrinkage reduces to pop-slope at $m{=}0$ ratings per user
(weight $\omega{=}0$) and overtakes the pop-slope baseline from
$\bm{m{=}5}$ ratings per user onward under random-fold CV, roughly a
four-fold improvement in data-efficiency since naive per-user OLS
only breaks even with pop-slope at $m{=}20$. The bias-variance
trade-off of
$\omega{=}\tau^2/(\tau^2+V)$ produces a non-monotone transition near
$m{=}3$, where shrinkage is worse than pop-slope on held-out RMSE.
We report this as a deployment-relevant failure mode; the operational
rule is to use pop-slope until $m{\ge}5$ ratings per user are
available, then switch to shrinkage.

\subsection{Ablations and failure cases}\label{sec:nulls}

We pre-registered four ablations, each tied to a specific claim it
could overturn; all four outcomes were consistent with the claims.
Together with three companion analyses they form seven stress
tests, summarized in three thematic
groups; the per-cell numerical detail is in
Appendix~\ref{app:prereg-mechanics}.

\begin{figure}[!b]
\centering
\includegraphics[width=0.74\columnwidth]{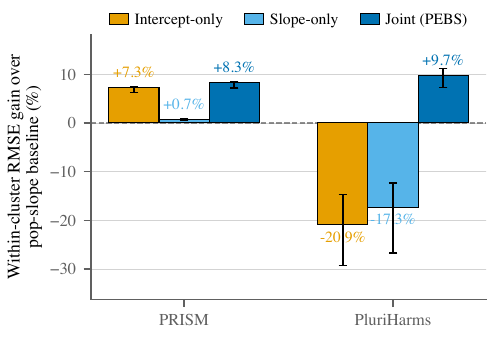}
\caption{\textbf{Both single-component estimators degrade RMSE on
PluriHarms; only the joint estimator yields the gain on both
corpora.} RMSE reduction (\%) vs.\ pop-slope, $95\%$ BCa CIs,
dashed reference at zero. Bars use the cross-corpus evaluation
protocol of Figure~\ref{fig:cross-corpus-forest}; the matched
single-corpus PRISM decomposition (\S\ref{sec:nulls}) agrees
within $0.25$\,pp.}
\label{fig:q3-half-arms}
\end{figure}

\textbf{(I) Mechanism necessity.} A leave-one-component-out
decomposition on PRISM shows that neither component suffices alone:
intercept-only shrinkage (the Efron--Morris floor) attains
${+}7.46\%$ and slope-only shrinkage ${+}0.74\%$, both strictly
below the joint gain; adding slope shrinkage on top of the intercept
floor contributes ${+}1.04$\,pp, and the slope component is the
only one that requires real RM signal, ruling out a pure-noise
explanation. The PluriHarms cross-corpus
replication (Figure~\ref{fig:q3-half-arms}) is the primary
evidence that intercept- and slope-shrinkage are jointly
necessary rather than additive: both single-component variants
(intercept-only and slope-only) degrade RMSE individually, yet the joint estimator is strictly dominant at
$\bm{+9.66\%}$. Method-of-Moments $\hat\tau^2_\alpha$ recovers the
ground-truth variance across the synthetic-seed grid; the sign-reversal and
adversarial-user injection probes both leave PEBS's RMSE below
the naive-no-pool baseline at every tested corruption level
(grid and per-cell numbers in the released artifact bundle).

\textbf{(II) Cross-axis generalization.} The pooled-multi-corpus
analysis is summarized in Figure~\ref{fig:cross-corpus-forest} and
the three-base-model and per-rater sample-efficiency analyses in
\S\ref{sec:stress-tests}; all return positive gain.
A three-base-model PRISM panel using mean-response log-likelihood
scoring~\citep{stiennon2020summarize} replicates the within-user
gain on each base model (per-cell numbers in the released
artifact bundle). Per-rater subsampling yields
monotone non-decreasing gain that tracks the Morris
$g$-function prediction $r/(1{+}r)$ across the swept sample budget.

\textbf{(III) Where PEBS does not improve.} Pair accuracy is
identical by construction across pop-slope and EB-shrunk arms
($0.6834$ in both arms on the CV evaluation pairs; this differs
from the $64.00\%$ in \S\ref{sec:setup}, which is the base RM's
held-out-user split), which
Proposition~\ref{prop:mi} predicts: PEBS
value lives in calibration-sensitive losses (RMSE, BT-NLL), not
argmax-style benchmarks like RewardBench~2. The HelpSteer2
verbosity attribute is the per-attribute null
($\omega_\beta\!\approx\!0.93$, gain straddles zero) while the
other four attributes gain positively;
the ordinal-preference limit on MultiPref is documented
separately in \S\ref{sec:morrisg}.

\section{Discussion}\label{sec:discussion}

\paragraph{Calibration vs.\ selection axes.}
Proposition~\ref{prop:mi} fixes pair accuracy across pop-slope and
EB-shrunk arms by construction; PEBS therefore targets
calibration-sensitive losses (\S\ref{sec:why-calib}) rather than
argmax-style benchmarks
(RewardBench~2~\citep{lambert2024rewardbench}). The
personalization gap that \citet{ma2026personalizedrb} report for
frontier RMs (peaking at $75.9\%$)
is not affected by PEBS's monotone calibration
(Proposition~\ref{prop:mi}); closing it requires a complementary selection-style
component. A complete per-rater system composes the upstream
RM, the \PEBS{} calibrator, and a selection-style component
such as ensemble disagreement~\citep{coste2024reward},
where PEBS targets calibration loss and the selection component
targets pair accuracy. A
complementary downstream procedure that \emph{relabels} preference
pairs using PEBS-corrected reward and trains a fresh DPO
policy~\citep{rafailov2023direct} falls outside
Proposition~\ref{prop:mi}'s scope: the relabeled training data
and the resulting policy's reward function both change. Across a
Llama-3-8B-Instruct base model and a Mistral-7B-Instruct-v0.3 with PRISM, this relabel-and-retrain
procedure yields $+8.81$\,pp on Llama-3-8B (single seed) and
$+12.90$\,pp on Mistral-7B-Instruct-v0.3 (seven-seed mean, $95\%$ CI
$[+11.97, +13.82]$, all seeds positive) held-out
pair accuracy on a user-disjoint $20\%$ held-out slice (i.e., $20\%$
of users not seen during training).

\paragraph{Pluralism at the individual rater scale.}
The \citet{sorensen2024roadmap} Roadmap distinguishes three
pluralism axes: distributional (a distribution over outputs),
steerable (conditionable on values or personas), and Overton (a
single output spanning the range). We propose calibration
heterogeneity as a fourth axis of pluralistic alignment: the
per-annotator slope $\alpha_j$ and offset $\beta_j$ at which each
rater converts a model score into a personal rating, which the
pooled-likelihood RLHF pipeline collapses into a single global
calibrator. \PEBS{} operates on this fourth axis. Distributional,
steerable, and Overton handle reasonable-disagreement-on-content;
calibration handles rater-specific scale-and-offset on a
fixed-content scoring task, and the four axes are complementary.
\PEBS{} retains per-rater $(\hat\alpha_j,\hat\beta_j)$ heterogeneity that
the pop-slope baseline pools away; the
\S\ref{sec:stress-tests} demographics-ANOVA null indicates that the
heterogeneity we recover is individual-rater and not
demographic-cohort. Cross-cultural value variation~\citep{conitzer2024socialchoice,zhang2025community}
is one plausible source of the residual $\hat\alpha_j,\hat\beta_j$
heterogeneity that demographics fail to explain; characterizing
the cultural-political content of these residuals is open work.
Shrinking per-rater calibrators toward a population
mean is a regression-to-mean operation, and the minority-rater
trade-off it entails is the subject of the next paragraph.

EB shrinkage with $\omega_j = \tau^2 / (\tau^2 + V_j)$ and
$V_j \propto 1/n_j$ shrinks low-$n_j$ raters more aggressively;
if those raters are disproportionately drawn from underrepresented
populations~\citep{kirk2024prism}, PEBS shrinks their estimated
$(\hat\alpha_j,\hat\beta_j)$ toward the population mean in exchange for variance reduction. This is the
standard shrinkage trade-off (Fig.~\ref{fig:f7-per-user}:
$71.9\%$ helped, $28.1\%$ hurt); when the rare-true-extreme tail
is policy-relevant a minority-rater audit is recommended.

\section{Limitations}\label{sec:limitations}

\paragraph{Base-family transfer scope.} The PEBS procedure
replicates within the Qwen-2.5 family across three corpora and
three base reward models (\S\ref{sec:stress-tests}) and on the
Phi-3-medium-14B same-family reference across $5$ training seeds
(all $5$ positive at coherence per-attribute mean $+42.15\%$;
Table~\ref{tab:base-transfer-grid}); the pre-registered four-base
coherence-only probe and verbosity-only control locate the
limit at a coherence-head/dense-architecture interaction rather
than verbosity bias, with calibration diagnostics in
Appendix~\ref{app:adapter-inspection}. On PRISM, $\hat\tau^2_\alpha$ is
dominated by between-rater value differences and not within-rater
rating noise (slope-SNR $15.6$ $[13.8, 17.5]$;
full residualization procedure in the released artifact bundle); the corresponding decomposition on
PluriHarms / OASST / SHP is not measured here. A regression-to-mean reading
of the control and a multi-seed verbosity-baseline probe
are open follow-ups.

\paragraph{Morris forecaster scope.} The Morris g-function
forecaster (\S\ref{sec:morrisg}) is validated here for
continuous-rating corpora; on MultiPref, its large
predicted-versus-observed gap flags that the ordinal preference
setting violates the Gaussian random-effects
assumption. Extending PEBS to ordinal data would require a
Beta-Binomial or Student-$t_\nu$ random-effects model.

\paragraph{Comparison to retained PRISM baselines.}
Against the four retained PRISM baselines on matched leave-one-conversation-out
(LOCO) PRISM, PEBS provides closed-form post-hoc calibration with no
test-time inference cost. P-GenRM exceeds PEBS under the same strict
LOCO RMSE protocol while using test-time prototype clustering; the
retained comparison is in App.~\ref{app:baseline-scope}
(Tab.~\ref{tab:method-axis}).

\paragraph{Data, licenses, and ethics.}
All preference corpora used in this work (PRISM~\citep{kirk2024prism},
PluriHarms~\citep{li2026pluriharms},
HelpSteer2~\citep{wang2024helpsteer2}, OASST2,
SHP-subreddit~\citep{ethayarajh2022understanding},
MultiPref~\citep{miranda2024hybrid}) are public datasets released by
their original authors under the licenses on the corresponding
dataset cards; this paper introduces no new human-subjects data
collection. The PEBS procedure produces only per-rater calibration
parameters; no rater identifiers are republished. Per-rater
shrinkage trades minority-rater $(\hat\alpha_j,\hat\beta_j)$ magnitude for variance
reduction (Fig.~\ref{fig:f7-per-user}); when policy decisions
hinge on the rare-true-extreme tail, a minority-rater audit is
recommended (\S\ref{sec:discussion}).

{\footnotesize
\bibliographystyle{icml2026}
\bibliography{references}

\begin{thebibliography}{48}
\providecommand{\natexlab}[1]{#1}
\providecommand{\url}[1]{\texttt{#1}}
\expandafter\ifx\csname urlstyle\endcsname\relax
  \providecommand{\doi}[1]{doi: #1}\else
  \providecommand{\doi}{doi: \begingroup \urlstyle{rm}\Url}\fi

\bibitem[Baker(2001)]{baker2001basics}
Baker, F.~B.
\newblock \emph{The Basics of Item Response Theory}.
\newblock ERIC Clearinghouse on Assessment and Evaluation, 2nd edition, 2001.

\bibitem[Bakker et~al.(2022)Bakker, Chadwick, Sheahan, Tessler,
  Campbell-Gillingham, Balaguer, McAleese, Glaese, Aslanides, Botvinick, and
  Summerfield]{bakker2022fineagree}
Bakker, M.~A., Chadwick, M.~J., Sheahan, H.~R., Tessler, M.~H.,
  Campbell-Gillingham, L., Balaguer, J., McAleese, N., Glaese, A., Aslanides,
  J., Botvinick, M.~M., and Summerfield, C.
\newblock Fine-tuning language models to find agreement among humans with
  diverse preferences.
\newblock In \emph{Advances in Neural Information Processing Systems
  (NeurIPS)}, 2022.

\bibitem[Bradley \& Terry(1952)Bradley and Terry]{bradley1952rank}
Bradley, R.~A. and Terry, M.~E.
\newblock Rank analysis of incomplete block designs: {I}. the method of paired
  comparisons.
\newblock \emph{Biometrika}, 39\penalty0 (3/4):\penalty0 324--345, 1952.

\bibitem[Cameron et~al.(2008)Cameron, Gelbach, and
  Miller]{cameron2008bootstrap}
Cameron, A.~C., Gelbach, J.~B., and Miller, D.~L.
\newblock Bootstrap-based improvements for inference with clustered errors.
\newblock \emph{The Review of Economics and Statistics}, 90\penalty0
  (3):\penalty0 414--427, 2008.

\bibitem[Castricato et~al.(2025)Castricato, Lile, Rafailov, Fr{\"a}nken, and
  Finn]{castricato2025persona}
Castricato, L., Lile, N., Rafailov, R., Fr{\"a}nken, J.-P., and Finn, C.
\newblock {PERSONA}: A reproducible testbed for pluralistic alignment.
\newblock In \emph{Proceedings of the 31st International Conference on
  Computational Linguistics (COLING)}, 2025.
\newblock arXiv:2407.17387.

\bibitem[Christiano et~al.(2017)Christiano, Leike, Brown, Martic, Legg, and
  Amodei]{christiano2017deep}
Christiano, P.~F., Leike, J., Brown, T.~B., Martic, M., Legg, S., and Amodei,
  D.
\newblock Deep reinforcement learning from human preferences.
\newblock In \emph{Advances in Neural Information Processing Systems
  (NeurIPS)}, 2017.

\bibitem[Conitzer et~al.(2024)Conitzer, Freedman, Heitzig, Holliday, Jacobs,
  Lambert, Moss{\'e}, Pacuit, Russell, Schoelkopf, Tewolde, and
  Zwicker]{conitzer2024socialchoice}
Conitzer, V., Freedman, R., Heitzig, J., Holliday, W.~H., Jacobs, B.~M.,
  Lambert, N., Moss{\'e}, M., Pacuit, E., Russell, S., Schoelkopf, H., Tewolde,
  E., and Zwicker, W.~S.
\newblock Position: Social choice should guide {AI} alignment in dealing with
  diverse human feedback.
\newblock In \emph{Proceedings of the 41st International Conference on Machine
  Learning (ICML)}, pp.\  9346--9360, 2024.
\newblock PMLR 235; arXiv:2404.10271.

\bibitem[Coste et~al.(2024)Coste, Anwar, Kirk, and Krueger]{coste2024reward}
Coste, T., Anwar, U., Kirk, R., and Krueger, D.
\newblock Reward model ensembles help mitigate overoptimization.
\newblock In \emph{International Conference on Learning Representations
  (ICLR)}, 2024.

\bibitem[Dawid \& Skene(1979)Dawid and Skene]{dawid1979maxlik}
Dawid, A.~P. and Skene, A.~M.
\newblock Maximum likelihood estimation of observer error-rates using the {EM}
  algorithm.
\newblock \emph{Journal of the Royal Statistical Society, Series C (Applied
  Statistics)}, 28\penalty0 (1):\penalty0 20--28, 1979.

\bibitem[Efron(1987)]{efron1987better}
Efron, B.
\newblock Better bootstrap confidence intervals.
\newblock \emph{Journal of the American Statistical Association}, 82\penalty0
  (397):\penalty0 171--185, 1987.

\bibitem[Efron \& Morris(1973)Efron and Morris]{efron1973stein}
Efron, B. and Morris, C.
\newblock Stein's estimation rule and its competitors--an empirical {B}ayes
  approach.
\newblock \emph{Journal of the American Statistical Association}, 68\penalty0
  (341):\penalty0 117--130, 1973.

\bibitem[Eisenstein et~al.(2024)Eisenstein, Nagpal, Agarwal, Beirami, D'Amour,
  Dvijotham, Fisch, Heller, Pfohl, Ramachandran, Shaw, and
  Berant]{eisenstein2024helping}
Eisenstein, J., Nagpal, C., Agarwal, A., Beirami, A., D'Amour, A., Dvijotham,
  D.~J., Fisch, A., Heller, K., Pfohl, S., Ramachandran, D., Shaw, P., and
  Berant, J.
\newblock Helping or herding? {R}eward-model ensembles mitigate but do not
  eliminate reward hacking.
\newblock In \emph{Conference on Language Modeling (CoLM)}, 2024.

\bibitem[Ethayarajh et~al.(2022)Ethayarajh, Choi, and
  Swayamdipta]{ethayarajh2022understanding}
Ethayarajh, K., Choi, Y., and Swayamdipta, S.
\newblock Understanding dataset difficulty with $\mathcal{V}$-usable
  information.
\newblock In \emph{International Conference on Machine Learning (ICML)}, 2022.

\bibitem[Gao et~al.(2023)Gao, Schulman, and Hilton]{gao2023scaling}
Gao, L., Schulman, J., and Hilton, J.
\newblock Scaling laws for reward model overoptimization.
\newblock In \emph{International Conference on Machine Learning (ICML)}, 2023.

\bibitem[Gelman \& Hill(2007)Gelman and Hill]{gelman2007data}
Gelman, A. and Hill, J.
\newblock \emph{Data Analysis Using Regression and Multilevel/Hierarchical
  Models}.
\newblock Cambridge University Press, 2007.

\bibitem[Ghafouri et~al.(2026)Ghafouri, Choi, Dey, and
  Ferrara]{ghafouri2026measuring}
Ghafouri, B., Choi, E.~C., Dey, P., and Ferrara, E.
\newblock Measuring human preferences in {RLHF} is a social science problem.
\newblock arXiv preprint arXiv:2604.03238, 2026.

\bibitem[Han et~al.(2026)Han, Zhou, Gao, Zhou, Li, Kumar, Fan, Li, and
  Zhang]{han2026ebpo}
Han, K., Zhou, Y., Gao, M., Zhou, G., Li, S., Kumar, A., Fan, X., Li, W., and
  Zhang, L.
\newblock {EBPO}: Empirical bayes shrinkage for stabilizing group-relative
  policy optimization.
\newblock arXiv preprint arXiv:2602.05165, 2026.

\bibitem[Henderson(1975)]{henderson1975best}
Henderson, C.~R.
\newblock Best linear unbiased estimation and prediction under a selection
  model.
\newblock \emph{Biometrics}, 31\penalty0 (2):\penalty0 423--447, 1975.

\bibitem[Hu et~al.(2022)Hu, Shen, Wallis, Allen-Zhu, Li, Wang, Wang, and
  Chen]{hu2022lora}
Hu, E.~J., Shen, Y., Wallis, P., Allen-Zhu, Z., Li, Y., Wang, S., Wang, L., and
  Chen, W.
\newblock {LoRA}: Low-rank adaptation of large language models.
\newblock In \emph{International Conference on Learning Representations
  (ICLR)}, 2022.

\bibitem[Kirk et~al.(2024)Kirk, Whitefield, R{\"o}ttger, Bean, Margatina, Ciro,
  Mosquera, Bartolo, Williams, He, Vidgen, and Hale]{kirk2024prism}
Kirk, H.~R., Whitefield, A., R{\"o}ttger, P., Bean, A., Margatina, K., Ciro,
  J., Mosquera, R., Bartolo, M., Williams, A., He, H., Vidgen, B., and Hale,
  S.~A.
\newblock The {PRISM} alignment dataset: What participatory, representative and
  individualised human feedback reveals about the subjective and multicultural
  alignment of large language models.
\newblock In \emph{Advances in Neural Information Processing Systems (NeurIPS,
  Datasets and Benchmarks Track)}, 2024.
\newblock arXiv:2404.16019.

\bibitem[Kobalczyk \& van~der Schaar(2025)Kobalczyk and van~der
  Schaar]{kobalczyk2025preference}
Kobalczyk, K. and van~der Schaar, M.
\newblock Preference learning for {AI} alignment: A causal perspective.
\newblock In \emph{International Conference on Machine Learning (ICML)}, 2025.
\newblock arXiv:2506.05967.

\bibitem[K{\"o}pf et~al.(2023)K{\"o}pf, Kilcher, von R{\"u}tte, Anagnostidis,
  Tam, Stevens, Barhoum, Duc, Stanley, Nagyfi, ES, Suri, Glushkov, Dantuluri,
  Maguire, Schuhmann, Nguyen, and Mattick]{kopf2023openassistant}
K{\"o}pf, A., Kilcher, Y., von R{\"u}tte, D., Anagnostidis, S., Tam, Z.-R.,
  Stevens, K., Barhoum, A., Duc, N.~M., Stanley, O., Nagyfi, R., ES, S., Suri,
  S., Glushkov, D., Dantuluri, A., Maguire, A., Schuhmann, C., Nguyen, H., and
  Mattick, A.
\newblock {OpenAssistant} conversations -- democratizing large language model
  alignment.
\newblock In \emph{Advances in Neural Information Processing Systems (NeurIPS,
  Datasets and Benchmarks Track)}, 2023.
\newblock arXiv:2304.07327.

\bibitem[Kou \& Yang(2017)Kou and Yang]{kou2017optimal}
Kou, S.~C. and Yang, J.~J.
\newblock Optimal shrinkage estimation in heteroscedastic hierarchical linear
  models.
\newblock In \emph{Big and Complex Data Analysis}, Contributions to Statistics,
  pp.\  249--284. Springer, 2017.
\newblock \doi{10.1007/978-3-319-41573-4_13}.

\bibitem[Li et~al.(2026)Li, Mire, Fleisig, Pyatkin, Collins, Sap, and
  Levine]{li2026pluriharms}
Li, J.-J., Mire, J., Fleisig, E., Pyatkin, V., Collins, A., Sap, M., and
  Levine, S.
\newblock {PluriHarms}: Benchmarking the full spectrum of human judgments on
  {AI} harm.
\newblock \emph{arXiv preprint arXiv:2601.08951}, 2026.

\bibitem[Liu et~al.(2024)Liu, Zeng, Liu, Yan, He, Wang, Yan, Liu, and
  Zhou]{liu2024skywork}
Liu, C.~Y., Zeng, L., Liu, J., Yan, R., He, J., Wang, C., Yan, S., Liu, Y., and
  Zhou, Y.
\newblock Skywork-reward: Bag of tricks for reward modeling in {LLMs}.
\newblock \emph{arXiv preprint arXiv:2410.18451}, 2024.

\bibitem[Liu et~al.(2025)Liu, Lu, and Sun]{liu2025uq}
Liu, P., Lu, J., and Sun, W.~W.
\newblock Uncertainty quantification for large language model reward learning
  under heterogeneous human feedback.
\newblock \emph{arXiv preprint arXiv:2512.03208}, 2025.

\bibitem[Ma et~al.(2026)Ma, Gao, Cai, Zhao, Zhou, Zhang, and
  Zhao]{ma2026personalizedrb}
Ma, Q., Gao, D., Cai, R., Zhao, B., Zhou, H., Zhang, J., and Zhao, Z.
\newblock Personalized {RewardBench}: Evaluating reward models with human
  aligned personalization.
\newblock arXiv preprint arXiv:2604.07343, 2026.

\bibitem[Malik et~al.(2025)Malik, Pyatkin, Land, Morrison, Smith, Hajishirzi,
  and Lambert]{lambert2024rewardbench}
Malik, S., Pyatkin, V., Land, S., Morrison, J., Smith, N.~A., Hajishirzi, H.,
  and Lambert, N.
\newblock {RewardBench 2}: Advancing reward model evaluation.
\newblock \emph{arXiv preprint arXiv:2506.01937}, 2025.

\bibitem[Miranda et~al.(2024)Miranda, Wang, Elazar, Kumar, Pyatkin, Brahman,
  Smith, Hajishirzi, and Dasigi]{miranda2024hybrid}
Miranda, L. J.~V., Wang, Y., Elazar, Y., Kumar, S., Pyatkin, V., Brahman, F.,
  Smith, N.~A., Hajishirzi, H., and Dasigi, P.
\newblock Hybrid preferences: Learning to route instances for human vs. {AI}
  feedback.
\newblock \emph{arXiv preprint arXiv:2410.19133}, 2024.

\bibitem[Morris(1983)]{morris1983parametric}
Morris, C.~N.
\newblock Parametric empirical {B}ayes inference: Theory and applications.
\newblock \emph{Journal of the American Statistical Association}, 78\penalty0
  (381):\penalty0 47--55, 1983.

\bibitem[Ouyang et~al.(2022)Ouyang, Wu, Jiang, Almeida, Wainwright, Mishkin,
  Zhang, Agarwal, Slama, Ray, Schulman, Hilton, Kelton, Miller, Simens, Askell,
  Welinder, Christiano, Leike, and Lowe]{ouyang2022training}
Ouyang, L., Wu, J., Jiang, X., Almeida, D., Wainwright, C.~L., Mishkin, P.,
  Zhang, C., Agarwal, S., Slama, K., Ray, A., Schulman, J., Hilton, J., Kelton,
  F., Miller, L., Simens, M., Askell, A., Welinder, P., Christiano, P., Leike,
  J., and Lowe, R.
\newblock Training language models to follow instructions with human feedback.
\newblock In \emph{Advances in Neural Information Processing Systems
  (NeurIPS)}, 2022.

\bibitem[Paun et~al.(2018)Paun, Carpenter, Chamberlain, Hovy, Kruschwitz, and
  Poesio]{paun2018comparing}
Paun, S., Carpenter, B., Chamberlain, J., Hovy, D., Kruschwitz, U., and Poesio,
  M.
\newblock Comparing bayesian models of annotation.
\newblock \emph{Transactions of the Association for Computational Linguistics
  (TACL)}, 6:\penalty0 571--585, 2018.

\bibitem[Pinheiro \& Bates(2000)Pinheiro and Bates]{pinheiro2000mixed}
Pinheiro, J.~C. and Bates, D.~M.
\newblock \emph{Mixed-Effects Models in {S} and {S}-{PLUS}}.
\newblock Statistics and Computing. Springer, 2000.

\bibitem[Rafailov et~al.(2023)Rafailov, Sharma, Mitchell, Manning, Ermon, and
  Finn]{rafailov2023direct}
Rafailov, R., Sharma, A., Mitchell, E., Manning, C.~D., Ermon, S., and Finn, C.
\newblock Direct preference optimization: Your language model is secretly a
  reward model.
\newblock In \emph{Advances in Neural Information Processing Systems
  (NeurIPS)}, 2023.

\bibitem[Rasch(1960)]{rasch1960probabilistic}
Rasch, G.
\newblock \emph{Probabilistic Models for Some Intelligence and Attainment
  Tests}.
\newblock Danmarks Paedagogiske Institut, Copenhagen, 1960.

\bibitem[Rezk et~al.(2025)Rezk, Pan, Foo, Xu, Chen, Gouk, and
  Hospedales]{rmselectioncrisis2025}
Rezk, F., Pan, Y., Foo, C.-S., Xu, X., Chen, N., Gouk, H., and Hospedales, T.
\newblock The reward model selection crisis in personalized alignment.
\newblock arXiv preprint arXiv:2512.23067, 2025.

\bibitem[Robbins(1956)]{robbins1956empirical}
Robbins, H.
\newblock An empirical {B}ayes approach to statistics.
\newblock In \emph{Proceedings of the Third Berkeley Symposium on Mathematical
  Statistics and Probability, Volume 1}, pp.\  157--163. University of
  California Press, 1956.

\bibitem[Schulman et~al.(2017)Schulman, Wolski, Dhariwal, Radford, and
  Klimov]{schulman2017proximal}
Schulman, J., Wolski, F., Dhariwal, P., Radford, A., and Klimov, O.
\newblock Proximal policy optimization algorithms.
\newblock \emph{arXiv preprint arXiv:1707.06347}, 2017.

\bibitem[Seabold \& Perktold(2010)Seabold and Perktold]{seabold2010statsmodels}
Seabold, S. and Perktold, J.
\newblock {statsmodels}: Econometric and statistical modeling with {Python}.
\newblock In \emph{9th Python in Science Conference (SciPy)}, 2010.

\bibitem[Sorensen et~al.(2024{\natexlab{a}})Sorensen, Jiang, Hwang, Levine,
  Pyatkin, West, Dziri, Lu, Rao, Bhagavatula, Sap, Tasioulas, and
  Choi]{sorensen2024kaleido}
Sorensen, T., Jiang, L., Hwang, J., Levine, S., Pyatkin, V., West, P., Dziri,
  N., Lu, X., Rao, K., Bhagavatula, C., Sap, M., Tasioulas, J., and Choi, Y.
\newblock Value kaleidoscope: Engaging {AI} with pluralistic human values,
  rights, and duties.
\newblock In \emph{AAAI Conference on Artificial Intelligence},
  2024{\natexlab{a}}.
\newblock arXiv:2309.00779.

\bibitem[Sorensen et~al.(2024{\natexlab{b}})Sorensen, Moore, Fisher, Gordon,
  Mireshghallah, Rytting, Ye, Jiang, Lu, Dziri, Althoff, and
  Choi]{sorensen2024roadmap}
Sorensen, T., Moore, J., Fisher, J., Gordon, M., Mireshghallah, N., Rytting,
  C.~M., Ye, A., Jiang, L., Lu, X., Dziri, N., Althoff, T., and Choi, Y.
\newblock Position: A roadmap to pluralistic alignment.
\newblock In \emph{International Conference on Machine Learning (ICML)},
  2024{\natexlab{b}}.
\newblock arXiv:2402.05070.

\bibitem[Stiennon et~al.(2020)Stiennon, Ouyang, Wu, Ziegler, Lowe, Voss,
  Radford, Amodei, and Christiano]{stiennon2020summarize}
Stiennon, N., Ouyang, L., Wu, J., Ziegler, D.~M., Lowe, R., Voss, C., Radford,
  A., Amodei, D., and Christiano, P.
\newblock Learning to summarize with human feedback.
\newblock In \emph{Advances in Neural Information Processing Systems
  (NeurIPS)}, 2020.

\bibitem[von Werra et~al.(2020)von Werra, Belkada, Tunstall, Beeching, Thrush,
  Lambert, Huang, Rasul, and Gallou{\'e}dec]{vonwerra2020trl}
von Werra, L., Belkada, Y., Tunstall, L., Beeching, E., Thrush, T., Lambert,
  N., Huang, S., Rasul, K., and Gallou{\'e}dec, Q.
\newblock {TRL}: Transformer reinforcement learning, 2020.
\newblock URL \url{https://github.com/huggingface/trl}.

\bibitem[Wang et~al.(2024)Wang, Dong, Delalleau, Zeng, Shen, Egert, Zhang,
  Sreedhar, and Kuchaiev]{wang2024helpsteer2}
Wang, Z., Dong, Y., Delalleau, O., Zeng, J., Shen, G., Egert, D., Zhang, J.~J.,
  Sreedhar, M.~N., and Kuchaiev, O.
\newblock {HelpSteer2}: Open-source dataset for training top-performing reward
  models.
\newblock \emph{arXiv preprint arXiv:2406.08673}, 2024.

\bibitem[Xie et~al.(2012)Xie, Kou, and Brown]{xie2012sure}
Xie, X., Kou, S.~C., and Brown, L.~D.
\newblock {SURE} estimates for a heteroscedastic hierarchical model.
\newblock \emph{Journal of the American Statistical Association}, 107\penalty0
  (500):\penalty0 1465--1479, 2012.

\bibitem[Yang et~al.(2025)Yang, Yang, Zhang, Hui, Zheng, Yu, Li, Liu, Huang,
  Wei, Lin, Yang, Tu, Zhang, Yang, Yang, Zhou, Lin, Dang, Lu, Bao, Yang, Yu,
  Li, Xue, Zhang, Zhu, Men, Lin, Li, Tang, Xia, Ren, Ren, Fan, Su, Zhang, Wan,
  Liu, Cui, Zhang, and Qiu]{qwen25}
Yang, A., Yang, B., Zhang, B., Hui, B., Zheng, B., Yu, B., Li, C., Liu, D.,
  Huang, F., Wei, H., Lin, H., Yang, J., Tu, J., Zhang, J., Yang, J., Yang, J.,
  Zhou, J., Lin, J., Dang, K., Lu, K., Bao, K., Yang, K., Yu, L., Li, M., Xue,
  M., Zhang, P., Zhu, Q., Men, R., Lin, R., Li, T., Tang, T., Xia, T., Ren, X.,
  Ren, X., Fan, Y., Su, Y., Zhang, Y., Wan, Y., Liu, Y., Cui, Z., Zhang, Z.,
  and Qiu, Z.
\newblock {Qwen2.5} technical report.
\newblock \emph{arXiv preprint arXiv:2412.15115}, 2025.

\bibitem[Zhang et~al.(2025)Zhang, Milli, Jusko, Smith, Amos, Bouaziz, Revel,
  Kussman, Sheynin, Titus, Radharapu, Yu, Sarma, Rose, and
  Nickel]{zhang2025community}
Zhang, L.~H., Milli, S., Jusko, K., Smith, J., Amos, B., Bouaziz, W., Revel,
  M., Kussman, J., Sheynin, Y., Titus, L., Radharapu, B., Yu, J., Sarma, V.,
  Rose, K., and Nickel, M.
\newblock Cultivating pluralism in algorithmic monoculture: The community
  alignment dataset.
\newblock arXiv preprint arXiv:2507.09650, 2025.

\bibitem[Zhang et~al.(2026)Zhang, Lin, Wu, Chen, Wang, Yang, Xu, Huang, Zhang,
  and Li]{pgenrm2026}
Zhang, P., Lin, T.-E., Wu, Y., Chen, J., Wang, Z., Yang, H., Xu, Z., Huang, F.,
  Zhang, K., and Li, Y.
\newblock {P-GenRM}: Personalized generative reward model with test-time
  user-based scaling.
\newblock In \emph{Proceedings of the International Conference on Learning
  Representations (ICLR)}, 2026.
\newblock Oral; arXiv:2602.12116.

\end{thebibliography}
}

\onecolumn
\appendix

\section{Proof of Theorem~\ref{thm:oracle} (oracle inequality)}
\label{app:proof-oracle}

We prove Theorem~\ref{thm:oracle} in four steps: (i) a mean-squared
error bound for the truncated Morris MoM estimator $\hat\tau^2$,
(ii) a second-order Taylor expansion with Lagrange remainder around
the oracle, (iii) aggregation across raters using the independence
delivered by sample splitting, and (iv) a truncation-event tail bound.

Throughout, $\tau^2$ abbreviates $\tau^2_\alpha$ and
$e_j = \hat\alpha_j^{\mathrm{OLS}}-\alpha_j$. Write
$\Delta_j(t)=\hat\alpha_j^{\mathrm{EB}}(t)-\alpha_j$
where $\hat\alpha_j^{\mathrm{EB}}(t)=\omega_j(t)\hat\alpha_j^{\mathrm{OLS}}+
(1-\omega_j(t))\alpha_{\mathrm{pop}}$ with
$\omega_j(t)=t/(t+V_j)$, so the per-rater risk at a deterministic $t$
is $R_{\mathrm{EB},j}(t)=\mathbb E[\Delta_j(t)^2]$ and the aggregate is
$R_{\mathrm{EB}}(t)=J^{-1}\sum_j R_{\mathrm{EB},j}(t)$. The oracle
risk is $R_{\mathrm{oracle}}=R_{\mathrm{EB}}(\tau^2)$.

\paragraph{Step 1: MoM mean-squared error.}
On the auxiliary split, \citet{morris1983parametric}'s estimator
\[
  \tilde\tau^2 = (J{-}1)^{-1}\!\sum_{j=1}^J
    (\hat\alpha_j^{\mathrm{OLS}}-\bar\alpha)^2 - J^{-1}\!\sum_{j=1}^J V_j
\]
is unbiased for $\tau^2$ under the random-effect DGP. Its first term
$S^2$ is a Gaussian quadratic form with matrix $A/(J-1)$,
$A = I - J^{-1}\mathbf 1\mathbf 1^{\top}$, applied to independent
coordinates of variance $\sigma_j^2 = \tau^2+V_j$, so with
$\Sigma = \mathrm{diag}(\sigma_j^2)$,
\[
  \mathrm{Var}(\tilde\tau^2)
  = \frac{2\,\mathrm{tr}\bigl[(A\Sigma)^2\bigr]}{(J-1)^2}
  \le \frac{2\sigma_{\max}^4}{J-1}
  \le \frac{8(\tau^2+V_{\max})^2}{3J} \equiv \frac{C_1}{J},
\]
using $\mathrm{tr}[(A\Sigma)^2]\le \|\Sigma\|^2\,\mathrm{tr}(A) =
\sigma_{\max}^4\,(J-1)$ and $1/(J-1)\le 4/(3J)$ for $J\ge 4$ (the only
place the $J\ge4$ assumption enters), giving
$\tilde\tau^2-\tau^2 = O_p(1/\sqrt J)$ and recovering
\citet{kou2017optimal}'s rate. Truncation at zero contracts the
squared error toward the truth when $\tau^2\ge 0$, so
$(\hat\tau^2-\tau^2)^2 \le (\tilde\tau^2-\tau^2)^2$ pointwise and
$\mathbb E[(\hat\tau^2-\tau^2)^2]\le C_1/J$, with
$C_1 = \tfrac83(\tau^2+V_{\max})^2 \le \tfrac83(1+M)^2\tau^4$.

\paragraph{Step 2: Taylor expansion.}
Fix rater $j$. For deterministic $t$, expanding $\Delta_j(t) =
\omega_j(t)\,e_j - (1-\omega_j(t))(\alpha_j - \alpha_{\mathrm{pop}})$
and using $\mathbb E[e_j^2]=V_j$,
$\mathbb E[(\alpha_j-\alpha_{\mathrm{pop}})^2]=\tau^2$, and
$\mathbb E[e_j(\alpha_j-\alpha_{\mathrm{pop}})]=0$ gives
\[
  R_{\mathrm{EB},j}(t) = \omega_j(t)^2 V_j + (1-\omega_j(t))^2 \tau^2 ,
\]
and $R_{\mathrm{EB},j}'(\tau^2)=0$: the oracle is a stationary point
of the per-rater risk. Hence the first-order term vanishes and a
second-order Taylor expansion with Lagrange remainder gives, for some
$\xi_j$ between $t$ and $\tau^2$,
\[
  R_{\mathrm{EB},j}(t)-R_{\mathrm{EB},j}(\tau^2)
   = \tfrac12 R_{\mathrm{EB},j}''(\xi_j)\,(t-\tau^2)^2.
\]
Direct computation gives
$R_{\mathrm{EB},j}''(\xi) = 2V_j^2(V_j+3\tau^2-2\xi)/(\xi+V_j)^4$,
which is decreasing in $\xi$ on $[\tau^2/2,\,3\tau^2/2]$. On the event
$\mathcal E=\{|\hat\tau^2-\tau^2|\le \tau^2/2\}$ we have
$\xi_j\in[\tau^2/2,\,3\tau^2/2]$ and therefore
\[
H_j \equiv \sup_{\xi\in[\tau^2/2,\,3\tau^2/2]} R_{\mathrm{EB},j}''(\xi)
 = R_{\mathrm{EB},j}''(\tau^2/2) \le \frac{64\,V_j^2}{(\tau^2+V_j)^3}.
\]

\paragraph{Step 3: Aggregation via sample splitting.}
Because $\hat\tau^2$ is computed on the auxiliary split, it is
independent of $\{e_j, \alpha_j\}$ for the $J$ raters being estimated,
so conditioning on $\hat\tau^2$ makes the deterministic-$t$ risk
formula of Step 2 applicable at $t=\hat\tau^2$:
\[
  \mathbb E\bigl[R_{\mathrm{EB}}(\hat\tau^2)
    -R_{\mathrm{oracle}}\,;\,\mathcal E\bigr]
  \;\le\; \tfrac{1}{2}\,\bar H\;
      \mathbb E\bigl[(\hat\tau^2-\tau^2)^2\bigr]
  \;\le\; \tfrac{1}{2}\,\bar H\,\frac{C_1}{J},
  \qquad \bar H = J^{-1}\!\sum_j H_j .
\]
Since $R_{\mathrm{oracle}} = J^{-1}\sum_j \tau^2 V_j/(\tau^2+V_j)$
and, for every $j$,
\[
\frac{64\,V_j^2/(\tau^2+V_j)^3}{\tau^2 V_j/(\tau^2+V_j)}
  = \frac{64\,V_j}{\tau^2(\tau^2+V_j)^2}
  \;\le\; \frac{16}{\tau^4}
\]
(the map $V \mapsto V/(\tau^2+V)^2$ is maximized at $V=\tau^2$), we
get $\bar H \le (16/\tau^4)\,R_{\mathrm{oracle}}$ and hence the
on-event bound
\[
  \mathbb E\bigl[R_{\mathrm{EB}}(\hat\tau^2)
    -R_{\mathrm{oracle}}\,;\,\mathcal E\bigr]
  \;\le\; \frac{c}{J}\,R_{\mathrm{oracle}},
  \qquad
  c = \frac{8\,C_1}{\tau^4} \le \frac{64}{3}(1+M)^2 .
\]

\paragraph{Step 4: Truncation event.}
Off $\mathcal E$, split the bad event into
$\mathcal A_-=\{\hat\tau^2 < \tau^2/2\}$ and
$\mathcal A_+=\{\hat\tau^2>3\tau^2/2\}$; these exhaust $\mathcal E^c$.
On $\mathcal A_-$, $R_{\mathrm{EB},j}(t)$ is decreasing on
$[0,\tau^2]$, so the per-rater risk is at most
$R_{\mathrm{EB},j}(0)=\tau^2$; on $\mathcal A_+$, $R_{\mathrm{EB},j}(t)$
is increasing on $[\tau^2,\infty)$ with limit $V_j$, so the per-rater
risk is at most $V_j \le M\tau^2$. The off-event excess risk is
therefore at most $\max(1,M)\,\tau^2$ per rater. For the probability,
$\tilde\tau^2-\tau^2$ is a centred Gaussian quadratic form whose
coefficient vector satisfies
$\|\lambda\|_\infty \le \sigma_{\max}^2/(J-1)$ and
$\|\lambda\|_2^2 \le \sigma_{\max}^4/(J-1)$, so the Hanson--Wright
inequality gives
\[
\mathbb P(\mathcal E^c)
 = \mathbb P\bigl(|\tilde\tau^2-\tau^2|>\tau^2/2\bigr)
 \le 2\exp\!\Bigl(-c_2\,\frac{J-1}{(1+M)^2}\Bigr)
\]
for an absolute constant $c_2>0$ (the exponent is dimensionless
because $\sigma_{\max}^2 \le (1+M)\tau^2$). Combining the on-event
Step~3 bound with the off-event excess and tail bound yields
Eq.~\eqref{eq:oracle-ineq}.\hfill$\square$

\paragraph{Scope of the proof.}
The theorem covers the sample-split estimator with
$\alpha_{\mathrm{pop}}$ known. Algorithm~\ref{alg:pebs} estimates
$\hat\tau^2$ and a precision-weighted $\hat\alpha_{\mathrm{pop}}$ on
the same sample; both couplings contribute additional $O(1/J)$ terms
(\citet{xie2012sure} handle the analogous same-sample coupling in the
location case via SURE). Rather than extending the algebra, we
validate the deployed same-sample estimator by simulation below. The
constant $\tfrac{64}{3}(1+M)^2$ is conservative: it is driven by the
sparsest raters through $V_{\max}$.

\paragraph{Empirical validation.}
We simulate the deployed (same-sample, truncated-MoM) estimator on
PRISM-calibrated cohorts: $J\in\{100, 200, 400, 800, 1394\}$ with
$100$ seeds each ($500$ cells), resampling $n_j$ from the empirical
PRISM pool with the fitted $\hat\tau^2{=}23.2$ and
$\hat\sigma_\varepsilon{=}23.5$. Define the realized constant
$c_{\mathrm{emp}} = J\,(R_{\mathrm{EB}}/R_{\mathrm{oracle}}-1)$.
Averaging risks over seeds within each stratum, the realized constant
is $3.93$ at $J{=}100$ and decreases to $2.51$ at $J{=}1394$; the
inequality holds in expectation in every stratum, with large slack
relative to the worst-case constant. Per-seed realized values
fluctuate with a heavy upper tail at small $J$ (95th percentile
${\approx}12$; single-seed maximum $91.6$ at $J{=}100$), as expected
for a ratio of noisy risk estimates; at $J{=}1394$ the mean risk
ratio is $1.002$ and the worst seed across $100$ is $1.017$. The
$\hat\tau^2$ estimation error is therefore too small to explain the
$8.58\%$ PRISM gain, which is consequently not an artefact of
estimating $\hat\tau^2$ from finite data.

\section{Additional diagnostics and pre-registration details}
\label{app:prereg-mechanics}
\label{app:adapter-inspection}
\label{app:cross-seed-per-attribute}

\begin{table}[ht]
\caption{\textbf{PRISM methods comparison.} Among the rows compared, PEBS is
the closed-form post-hoc calibrator with no test-time compute and
a stated oracle bound. P-GenRM is included as the matched scalar-RMSE
baseline and is evaluated under the strict LOCO protocol.}
\label{tab:method-axis}
\centering\scriptsize
\setlength{\tabcolsep}{2.5pt}
\renewcommand{\arraystretch}{1.05}
\begin{tabular}{@{}p{0.24\textwidth}p{0.18\textwidth}p{0.25\textwidth}p{0.25\textwidth}@{}}
\toprule
Method & RMSE gain $\uparrow$ & Test-time compute $\downarrow$ & Notes \\
\midrule
\textbf{Pop-slope baseline} & $0$ (ref.) & none & pooled affine calibrator \\
\textbf{Naive per-user OLS} & $+7.02\%$ & none & high-variance at low $n_j$ \\
Efron--Morris intercept-only EB & $+7.46\%$; $+4.70\%$ LOCO & none & intercept shrinkage only \\
\textbf{PEBS (ours)} & $\mathbf{+8.58\%}$; $+5.88\%$ LOCO & \textbf{none} & stable per-user IDs; Theorem~\ref{thm:oracle} \\
\midrule
P-GenRM~\citep{pgenrm2026} & $+8.13\%$ LOCO & prototype clustering + per-prototype OLS & learned prototypes \\
\bottomrule
\end{tabular}
\end{table}

\begin{figure}[ht]
\centering
\begin{minipage}{0.48\textwidth}
\centering
\includegraphics[width=\textwidth]{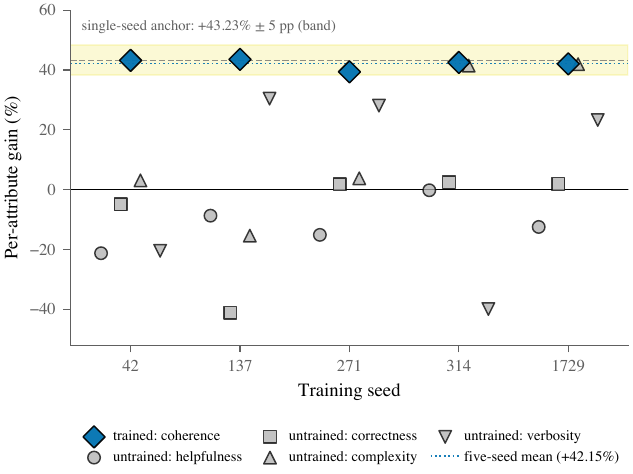}
\caption{\textbf{Phi-3-medium-14B cross-seed scatter.} Each column is
one random seed, and each colour-coded marker series is one HelpSteer2
attribute. Trained-attribute coherence is positive in all five
seeds (mean $+42.15\%$, dotted line; Student-$t$ $95\%$ CI
$[+40.10,+44.20]$); the shaded band marks the single-seed anchor
$+43.23\% \pm 5$\,pp. Untrained attributes scatter more widely.}
\label{fig:f19-prime-scatter}
\end{minipage}\hfill
\begin{minipage}{0.48\textwidth}
\centering
\includegraphics[width=\textwidth]{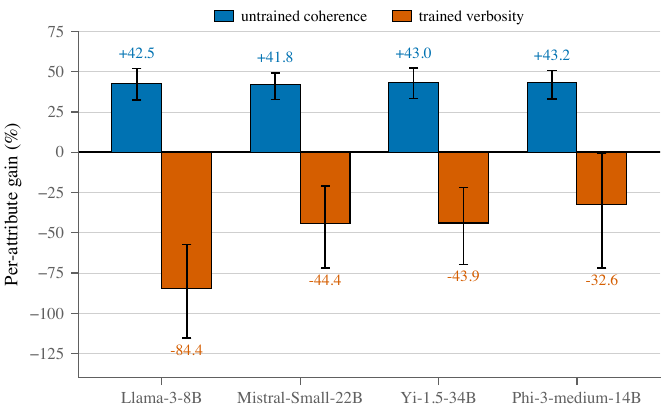}
\caption{\textbf{The verbosity-only control confirms the reversal
is attribute-specific, not architecture-wide.} One group per base:
blue bars are the untrained-coherence gain (\%), vermillion bars the
trained-verbosity gain (\%), with $95\%$ BCa CIs. On all four bases
the untrained coherence head stays positive and within
${\sim}1$\,pp of the Phi-3 reference, while the trained verbosity
head itself turns negative ($-84.4$ / $-44.4$ / $-43.9$ / $-32.6$),
ruling out a base-level failure as the cause of the coherence
reversal.}
\label{fig:verbosity-null-panel}
\end{minipage}
\end{figure}

This appendix expands the diagnostics supporting
Figure~\ref{fig:overview}: sparse-rater shrinkage, cross-base
transfer boundaries, adapter prediction-spread, and the numerical
details needed to reproduce the reported CIs.

\begin{figure}[ht]
\centering
\includegraphics[width=0.55\textwidth]{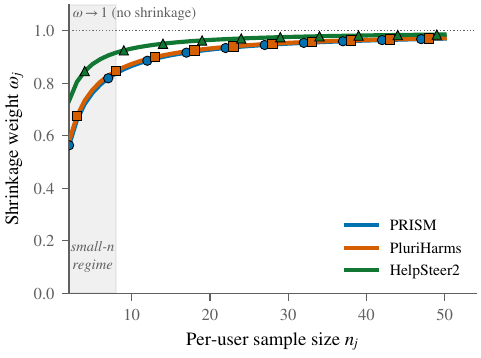}
\caption{\textbf{PEBS automatically down-weights sparse annotators:
shrinkage is largest for $n_j{\le}8$ and fades to zero at high $n_j$
with no threshold to tune.} The closed-form weight
$\omega_j{=}\tau^2/(\tau^2{+}V_j)$ governs how much PEBS trusts each
rater's own calibrator vs.\ the population mean. Three illustrative
populations (PRISM, PluriHarms, HelpSteer2) are plotted against per-user
sample sizes $n_j$ and within-rater noise $V_j\propto 1/n_j$; the
shaded zone ($n_j{\le}8$) is where $\omega_j$ is smallest and shrinkage
toward $\alpha_{\mathrm{pop}}$ is largest.
$\omega_j$ asymptotes to $1$ (no shrinkage) as $n_j$ grows, so dense
annotators reduce to per-user OLS without any threshold parameter.}
\label{fig:omega-shrinkage}
\end{figure}

\paragraph{Scope of the pre-registered criterion.}
The four-base coherence-only probe was pre-registered with the
criterion that any single base inversion bounds the across-family
result. The dense panel therefore supports the bounded-transfer
claim, not an architecture-universal claim. Two MoE runs
(Phi-3.5-MoE-Instruct and Mixtral-$8\!\times\!7$B) used a
narrower output-projection adapter than the dense-Transformer
protocol; both produce negative-direction trained-coherence gains
($-59.41\%$ and $-60.44\%$) with narrow prediction spread
($0.267$ and $0.298$). We use these MoE points only as additional
boundary evidence consistent with the dense-panel collapse
signature, not as full cross-architecture replications.

\begin{figure}[ht]
\centering
\includegraphics[width=0.72\columnwidth]{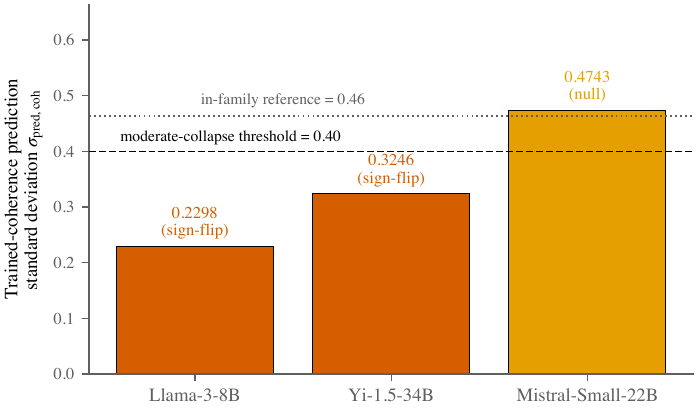}
\caption{\textbf{Adapter prediction-spread ($\sigma_{\mathrm{pred,coh}}$)
for three across-family bases, against the 0.40 collapse threshold.}
The two inversion bases (Llama-3-8B, Yi-1.5-34B) fall below the
threshold; the null base (Mistral-Small-22B) lies above.
Lower $\sigma_{\mathrm{pred,coh}}$ indicates tighter clustering of
adapter outputs around one or two rating values, consistent with
a head-collapsed adapter setting. We treat the signature as
observational rather than causal.}
\label{fig:head-collapse-signature}
\end{figure}

\paragraph{Calibration diagnostics.} The probe measures prediction
spread on a $208$-row HelpSteer2 slice. The two inversion bases
(Llama-3-8B, Yi-1.5-34B) lie below the $0.40$ collapse threshold
($\sigma_{\text{pred,coh}}{=}0.2298, 0.3246$), while
Mistral-Small-22B lies above ($0.4743$);
Figure~\ref{fig:head-collapse-signature} plots the three values.
Verbosity-bias and LoRA-capacity alternatives are addressed by
the verbosity-only control and the multi-seed Phi-3 replication. We
treat head collapse as an observational signature: the posterior
is wide ($P\!\in\![0.30,\,0.85]$), and causal mechanism claims
require intervention experiments outside this paper.

\begin{figure}[ht]
\centering
\includegraphics[width=0.85\columnwidth]{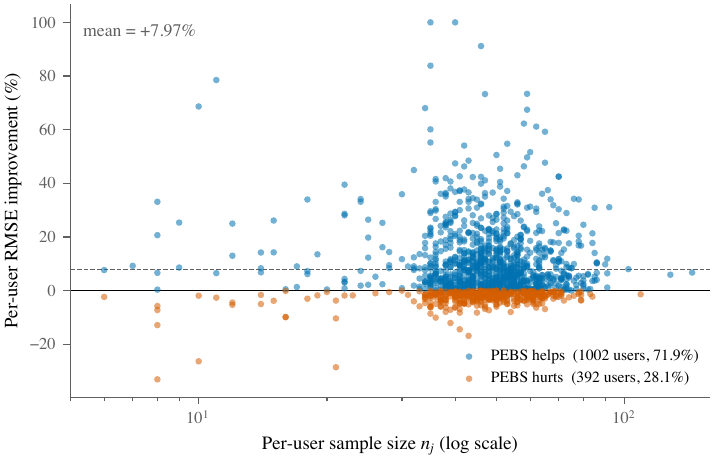}
\caption{\textbf{Per-user RMSE improvement scatter on PRISM ($N{=}1{,}394$
users), illustrating the minority-rater trade-off of EB shrinkage.}
Each point is one user; the $x$-axis is per-user sample size $n_j$
(log scale) and the $y$-axis is the per-user RMSE improvement
(pop-slope minus PEBS-shrunk, as a percentage of pop-slope RMSE).
Blue points ($1{,}002$ users, $71.9\%$) are helped by PEBS;
vermillion points ($392$, $28.1\%$) are hurt. Low-$n_j$ users are shrunk most aggressively
($\omega_j \to 0$ as $n_j \to 0$) and show the widest spread in
improvement, consistent with the standard EB trade-off: optimal under
the prior but wrong for the rare true-extreme rater.}
\label{fig:f7-per-user}
\end{figure}

\paragraph{Demographic ANOVA on PRISM.}
An Analysis of Variance of the fitted $(\hat\alpha_j,\hat\beta_j)$
against the same six PRISM demographics (age, gender, region, education,
political orientation, English fluency) finds only the
gender${\to}\hat\beta_j$ cell surviving Bonferroni correction,
and even there the explained variance is small ($\eta^2{<}0.02$).
Demographic grouping cannot replace per-user calibration; the
six demographic axes do not jointly recover the per-user shrinkage
gain reported in \S\ref{sec:withinuser}.

\paragraph{Multi-attribute regression observation on HelpSteer2.}
\label{app:helpsteer-multi-attribute}
We also observe the same shrinkage mechanism on a
multi-attribute regression problem, where the five HelpSteer2
attribute axes are treated as five pseudo-raters across $1{,}038$
rows. This is an observation about EB-shrinkage stability in a
multi-axis regression context, \emph{not} a pluralism claim: the
five axes are scoring dimensions, not human annotators with
heterogeneous calibrations. The four-seed Qwen-2.5-7B mean is
${+}18.24\%$ relative RMSE reduction $[+17.97, +18.51]$
(across-seed half-width $0.27$\,pp), reflecting the same
calibration-loss-reduction PEBS provides on PRISM applied to a
different problem geometry. The result is bound to the Qwen-2.5
family (\S\ref{sec:crossfam}, \S\ref{sec:limitations}); the
across-seed half-width is roughly an order of magnitude tighter
than the within-seed bootstrap half-width.

\paragraph{HelpSteer2 verbosity per-attribute null.}
Among the five HelpSteer2 attributes the EB-shrunk arm gains
positively on four (helpfulness ${+}6.10\%$, correctness
${+}7.08\%$, coherence ${+}41.15\%$, complexity ${+}30.13\%$);
verbosity straddles zero at ${-}2.74\%$ $[-38.04,\,{+}27.62]$
with shrinkage weight $\omega_\beta\!\approx\!0.93$, indicating
the attribute is already near-saturated under per-attribute fit
and there is little for shrinkage to add. The four-of-five
positive pattern rules out an attribute-agnostic verbosity bias
as the source of the within-user RMSE gain in \S\ref{sec:withinuser}.

\paragraph{Base-model training details.}
\label{app:replication} The five-seed Phi-3 replication gives
cross-seed mean ${+}42.15\%$, within $1.08$\,pp of the single-seed
reference (${+}43.23\%$), with trained-coherence across-seed
variance $2.73$\,pp$^2$ (SD $1.65$\,pp) versus
untrained mean $580.8$\,pp$^2$. Phi-3 verbosity-only
control turns trained-verbosity negative to ${-}32.62\%$ while
preserving untrained-coherence at ${+}43.18\%$
(Table~\ref{tab:base-transfer-grid}). Qwen2.5-7B-Instruct uses
Transformer Reinforcement Learning (TRL) 0.12.2~\citep{vonwerra2020trl} LoRA $r{=}32$, $\alpha{=}16$,
lr $10^{-4}$, bf16, $1{,}500$ steps, centered-rewards
regularizer~\citep{eisenstein2024helping}, pair accuracy CI
$[62.74, 65.29]$, $\approx 75$\,min H100 80\,GB. Bootstrap CIs
are $95\%$ BCa~\citep{efron1987better} with a PRISM
$4{,}000$-replicate cluster bootstrap by
user~\citep{cameron2008bootstrap} and a HelpSteer2 row-cluster.
PRISM MoM: $\hat\tau_\alpha^2{=}26.2$ (slope),
$\hat\tau_\beta^2{=}115.7$ (offset), $\hat\sigma_\varepsilon{=}23.5$ (residual
SD of the population-calibrator fit; per-user calibration takes
held-out RMSE below this value, Table~\ref{tab:t1-within}).
The $8.58\%$ random-fold-within-user PRISM result attenuates
predictably under stricter splits: a strict temporal $80/20$
returns $+7.55\%$ ($30$-seed cluster-bootstrap CI
$[+6.82, +8.71]$), cluster-bootstrap-by-user gives $+6.96\%$
(BCa $[+6.40, +7.56]$), and leave-one-conversation-out yields
$+5.88\%$ (BCa $[+5.17, +6.63]$); all four exclude zero.

\section{PRISM baseline scope}
\label{app:baseline-scope}

P-GenRM~\citep{pgenrm2026} is included as the matched scalar-RMSE
baseline and exceeds \PEBS{} in the strict LOCO cell reported in
Table~\ref{tab:method-axis}.
Methods whose published protocols optimise a different objective,
metric, or feature space are cited in related work but are not
reproduced as direct scalar-RMSE comparison rows here, since the protocol
mismatch makes the resulting numbers incomparable.

\section{Dataset cards}
\label{app:datasets}

This appendix expands the corpora used in
\S\ref{sec:4corpora} (the three within-scope continuous-rating
corpora) and \S\ref{sec:morrisg} (MultiPref, the
theory-predicted scope-limit demonstration corpus), with details
on collection, structure, and the operations PEBS requires. None
of these corpora is collected by us.

\paragraph{PRISM Alignment corpus~\citep{kirk2024prism}.}
A public preference-elicitation corpus with 1{,}500 unique
participants drawn from 75 countries and 24 demographic axes.
Each participant has a stable per-annotator ID and contributes
multi-turn conversations with multiple model variants, with both
turn-level (Likert 0--100) ratings and pairwise preferences. PRISM
is the primary evaluation corpus for PEBS because the per-annotator
IDs are stable across conversations, which is required to estimate the
per-user $(\alpha_j, \beta_j)$ random effect. The reward model is trained on $26{,}876$ preference pairs from
the $1{,}391$ demographic-complete participants under an $80/20$
stratified-by-user split; the per-rater calibrators use the
$1{,}394$-user utterance-level cohort ($n_j \ge 6$; \S\ref{sec:setup}).

\paragraph{PluriHarms~\citep{li2026pluriharms}.}
A harm-rating corpus collecting $15{,}000$ harm ratings on a
$0$--$100$ scale from $100$ annotators across $150$ prompts. Each
prompt-response pair is rated by multiple annotators with a stable
per-annotator ID. PluriHarms tests whether the PEBS procedure transfers
from preference judgments (PRISM) to a qualitatively different
feedback type (single-axis harm rating).

\paragraph{MultiPref~\citep{miranda2024hybrid}.}
A five-point Likert preference corpus in which annotators express
preferences with confidence ratings rather than as binary
BT-style picks. The per-annotator rating
distribution is non-Gaussian, so MultiPref lies outside the
Gaussian random-effects regime that PEBS's MoM estimator
assumes. The corpus enters this paper only as the
theory-predicted scope-limit demonstration discussed in
\S\ref{sec:morrisg}; the negative-control framing, the
predicted-versus-observed numerical gap, and the principled
Beta-Binomial or Student-$t_\nu$ random-effects extension are
all documented there.

\paragraph{HelpSteer2 attribute-as-rater recast~\citep{wang2024helpsteer2}.}
HelpSteer2 provides five scalar attribute ratings per prompt-response
pair (helpfulness, correctness, coherence, complexity, verbosity) on
a $0$--$4$ scale, from a panel of human annotators whose individual
identities are not released. Because PEBS requires per-rater data,
we re-cast the corpus by treating the five attribute axes themselves
as five \emph{pseudo-raters}: each row contributes one rating from
each axis, so a single prompt-response pair is rated by all five
attribute ``raters''. The HelpSteer2 attribute-as-rater protocol uses
$1{,}038$ rows. The cross-family probes of \S\ref{sec:crossfam}
train a coherence-only LoRA adapter (loss masked to the coherence
axis) and a verbosity-only counterfactual (loss masked to verbosity)
on each of the four pre-registered base architectures (plus the two
appendix-only MoE boundary runs).

\paragraph{Forecast companion corpora.}
OASST2-author~\citep{kopf2023openassistant} and
SHP-subreddit~\citep{ethayarajh2022understanding} are open
preference corpora with stable author- or subreddit-level grouping
variables. OASST2-author enters the paper twice, under two distinct
protocols: the \S\ref{sec:4corpora} within-cluster replication
(model-likelihood covariate, $1{,}017$ authors at $n_j{\ge}6$,
$5$-fold CV with cluster bootstrap; ${+}1.21\%$) and the
\S\ref{sec:morrisg} forecaster validation (rank covariate,
$2{,}507$ authors at $n_j{\ge}5$, leave-one-row-out CV;
$8.33\%$). SHP-subreddit enters only the forecaster validation
($18$ subreddit clusters at $n{\ge}20$); at these cluster sizes the
shrinkage weight saturates ($\omega\!\to\!1$), PEBS reduces to
per-cluster OLS, and the exact $0.00$\,pp forecast match is
expected rather than informative.

\end{document}